\documentclass[10pt,twocolumn,letterpaper]{article}

\usepackage[pagenumbers]{cvpr} 

\newcommand{\fakeref}[1]{{\textcolor{blue}{}}}

\usepackage{graphicx}
\usepackage{amsmath}
\usepackage{amssymb}
\usepackage{booktabs}
\usepackage{microtype}
\usepackage{enumitem}
\usepackage{array,multirow,textcomp}
\usepackage{balance}
\usepackage[normalem]{ulem}

\usepackage{macros}
\usepackage{shortbold}
\definecolor{cvprblue}{rgb}{0.21,0.49,0.74}
\usepackage[pagebackref,breaklinks,colorlinks,allcolors=cvprblue]{hyperref}
\usepackage{comment}
\usepackage[capitalize]{cleveref}
\crefname{section}{Sec.}{Secs.}
\Crefname{section}{Section}{Sections}
\Crefname{table}{Table}{Tables}
\crefname{table}{Tab.}{Tabs.}
\Crefname{equation}{Equation}{Equations}
\crefname{equation}{Eq.}{Eqs.}
\Crefname{figure}{Figure}{Figures}
\crefname{figure}{Fig.}{Figs.}
\Crefname{table}{Table}{Tables}
\crefname{table}{Tab.}{Tabs.}

\def\paperID{1234} 
\def\confName{CVPR}
\def\confYear{2025}

\title{SIR-DIFF: Sparse Image Sets Restoration with Multi-View Diffusion Model}

\author{
Yucheng Mao\textsuperscript{*} \quad
Boyang Wang\textsuperscript{*} \quad
Nilesh Kulkarni \quad
Jeong Joon Park
\vspace{5mm} \\
University of Michigan, Ann Arbor \quad 
}

\begin{document}

\twocolumn[{
\renewcommand\twocolumn[1][]{#1}
\maketitle

\begin{center}
    \vspace{-0.4cm}
    \centering
    \includegraphics[width=0.9\linewidth]{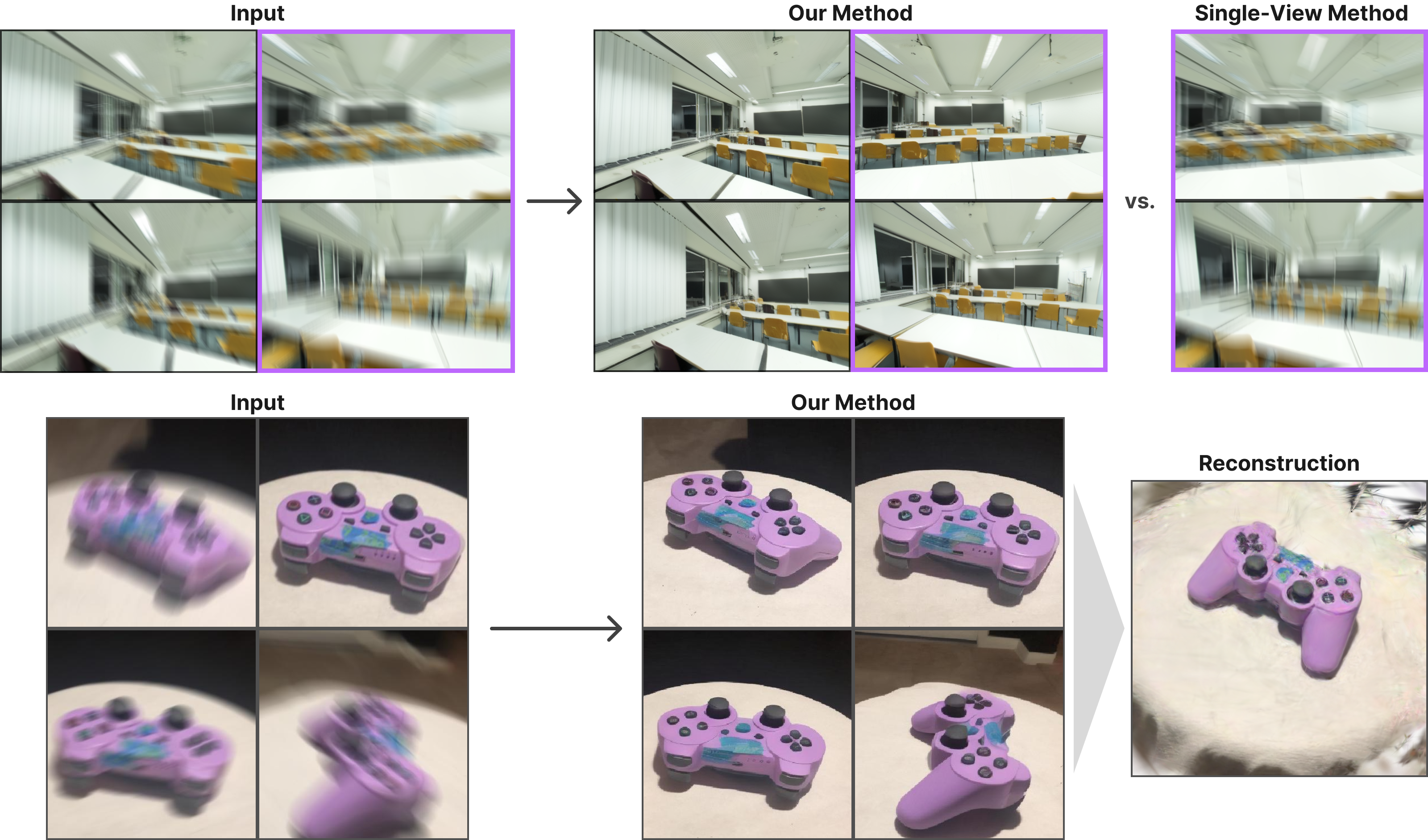}
    \vspace{-0.1cm}
    \captionof{figure}{
    \textbf{Sparse View Image Restoration}. Our diffusion model takes multi-view images and jointly enhances their visual quality while maintaining 3D consistency. (Top) Four motion-blurred input images are processed by our method, resulting in sharp outputs that significantly outperform single-view restoration methods as shown in the corresponding purple boxes. (Bottom) Our method can consistently restore multi-view images (4 out of 50 shown), leading to accurate 3D reconstructions.
    }
    \vspace{-0.cm}
\label{fig:teaser}
\end{center}

}]

\maketitle
{\let\thefootnote\relax\footnotetext{{{*} Equal contribution}}}
\vspace{-0.1cm}
{\centering\large\bfseries Abstract\par}
\vspace{-0.0cm}
{\itshape
The computer vision community has developed numerous techniques for digitally restoring true scene information from single-view degraded photographs, an important yet extremely ill-posed task. In this work, we tackle image restoration from a different perspective by jointly denoising multiple photographs of the same scene. Our core hypothesis is that degraded images capturing a shared scene contain complementary information that, when combined, better constrains the restoration problem. To this end, we implement a powerful multi-view diffusion model that jointly generates uncorrupted views by extracting rich information from multi-view relationships. Our experiments show that our multi-view approach outperforms existing single-view image and even video-based methods on image deblurring and super-resolution tasks. Critically, our model is trained to output 3D consistent images, making it a promising tool for applications requiring robust multi-view integration, such as 3D reconstruction or pose estimation. Project website: \url{https://myc634.github.io/sirdiff/}\\
}
    
\vspace{-0.5cm}
\section{Introduction}
\vspace{-0.2cm}

Image restoration approaches~\cite{wang2021real, dong2015compression, ho2020denoising, ma2022deblur} typically process degraded images in isolation, confronting the challenge of the original scene content. 
Yet in practice, we frequently capture multiple photographs of meaningful moments, yielding a natural collection of views of the same scene, albeit some degraded. This common behavior presents an untapped opportunity: leveraging multiple captures to enhance image quality and better preserve our visual memories. 

In this work, we present a novel perspective on image restoration by reformulating it as a multi-view collaborative restoration task. 
We leverage our key insight that multiple corrupted photographs of the same scene contain complementary information that can aid superior denoising outcomes. 
The concept of multi-image fusion for quality enhancement has been explored in classical super-resolution techniques where sub-pixel shifted images are combined to reconstruct fractional pixel information~\cite{kawulok2019deep}. Our approach revisits this paradigm in a fundamentally wider context by using neural attention mechanisms to implicitly fuse overlapping information from general multi-view images to enhance even severe aberrations.

Notably, our multi-view restoration framework has particular significance for 3D computer vision applications. Many real-world tasks, from robotic SLAM to novel-view synthesis, rely fundamentally on multi-view imagery of a scene. When restoring degraded images within such multi-view sets, maintaining geometric consistency across views is crucial to preserve the underlying 3D scene assumptions. Traditional single-view restoration methods, inevitably introduce inconsistencies that can compromise these geometric constraints. In contrast, our restoration approach naturally enforces view consistency through learned neural priors.

We implement our approach, dubbed SIR-Diff ({\bf S}parse {\bf I}mage {\bf R}estoration with {\bf Diff}usion), through an innovative multi-view diffusion model architecture. Our choice of diffusion models is motivated by their natural handling of uncertainty in image restoration. We extend the UNet-based Stable Diffusion~\cite{rombach2022high} architecture to process multiple views jointly by introducing three key components: (1) a unified encoder for diverse degraded inputs including low-resolution, blurred, and regular RGB images, (2) a hybrid 2D-3D convolutional structure derived from SVD (Stable Video Diffusion)~\cite{blattmann2023align, blattmann2023stable}, and (3) a 3D self-attention Transformer that enables cross-view attention across all spatial and multi-view tokens.

We evaluate and show zero-shot performance on Scannet++~\cite{yeshwanth2023scannet++}, ETH3D~\cite{schops2019bad}, and CO3D~\cite{reizenstein2021common} for multi-view deblurring and super-resolution tasks and their 3D downstream applications. \ours consistently outperforms existing single-image and video restoration methods for multi-view restoration tasks in both image quality and our proposed 3D view consistency metrics.  For downstream applications, we show that our multi-view restoration strategy significantly enhances 3D reconstruction quality, as evidenced by improved novel-view synthesis accuracy and feature-matching success rates. Beyond image restoration, our architecture shows promise in other multi-view tasks, notably achieving competitive depth estimation accuracy and geometric consistency compared to leading single-view methods.

\label{sec:intro}

\section{Related Works}
\label{sec:related_works}

\parnobf{Image Diffusion Models} Diffusion Models (DM) show huge popularity in all domains: illustrations~\cite{zhang2023adding, wang2024instantid, esser2024scaling, rombach2022high}, video creation~\cite{blattmann2023stable, polyak2024movie, blattmann2023align}, 3D modeling with limited input views~\cite{long2024wonder3d, shi2023mvdream, voleti2025sv3d}, 4D generation~\cite{ling2024align, bahmani20244d, bahmani2025tc4d},  
and \etc.
Most models are fine-tuned from the pre-trained weight instead of training from the beginning.
One of the most widely-used pre-trained Diffusion Models is Stable Diffusion 2.1~\cite{rombach2022high}, a UNet-based~\cite{ronneberger2015u} architecture.

\parnobf{Multiview-Image Diffusion models} In the image-to-3D task~\cite{li2023generative}, Diffusion Model is a powerful generative prior to enriching details for the novel (unseen) view synthesis~\cite{hollein2023text2room, chan2023generative, watson2022novel, wang2024prolificdreamer, tewari2023diffusion}.
Methods like MVDream~\cite{shi2023mvdream}, CAT3D~\cite{gao2024cat3d}, Zero123++~\cite{shi2023zero123++}
jointly generate multi-views simultaneously by extending the image diffusion model (SD2.1) with a 3D self-attention layer. However, these works ignore the exploration of the convolution layer architecture in the UNet-based Diffusion Model. In this paper, we explore the methodology of applying the 3D self-attention layer to the 3D restoration task.

\parnobf{Image and Video Restoration} These methods aim at converting a degraded input image or video into a high-quality form.
Traditional tasks include Super-Resolution~\cite{ledig2017photo, wang2021real, yu2024scaling, wu2024one, zhou2024upscale}, Deblurring~\cite{liang2024vrt, kong2023efficient, potlapalli2024promptir}, Denoise~\cite{yang2023real, fan2019brief}, Compression Artifact Reduction~\cite{dong2015compression, wang2024vcisr, wang2024apisr}, Inpainting~\cite{chung2022diffusion, xiang2023deep}, Burst Image Deblurring~\cite{aittala2018burst, guo2022differentiable}, and \etc.
These domains have achieved milestones in the 2D image and video domain~\cite{luo2024taming}. 
However, the study on the restoration in 3D is insufficient. 
The restoration consistency of the texture and geometry from different views is hard to maintain when independently restoring them.

\parnobf{3D Reconstruction from Corrupted Inputs}
3D Gaussian Splatting (3DGS)~\cite{kerbl20233d, wang2025view, ren2023dreamgaussian4d, yu2024gsdf} transforms input images into explicit-represented 3D Gaussian blobs with continuous optimization to reconstruct the large 3D scene.
However, these models heavily rely on the condition of the input images. 
Motion blurring~\cite{wang2023bad, zhao2024bad} or low-resolution~\cite{shen2024supergaussian, feng2024srgs} inputs that lack high-frequency and sharp details may greatly deteriorate the reconstruction quality or even fail at the initialization stage.
To resolve this, methods like Bad-Gaussian~\cite{shen2024supergaussian} and SuperGaussian~\cite{zhao2024bad} depend on dense view input to iteratively optimize the degraded artifacts. Further, under heavily deteriorated inputs, these methods cannot recover high-quality outputs when they lack a strong generative prior~\cite{shen2024supergaussian} such as a pre-trained Diffusion Model.

\section{Method}
\label{sec:method}

\begin{figure*}[t]
    \centering
    \vspace{-0.3cm}
    \includegraphics[width=1.0\linewidth]{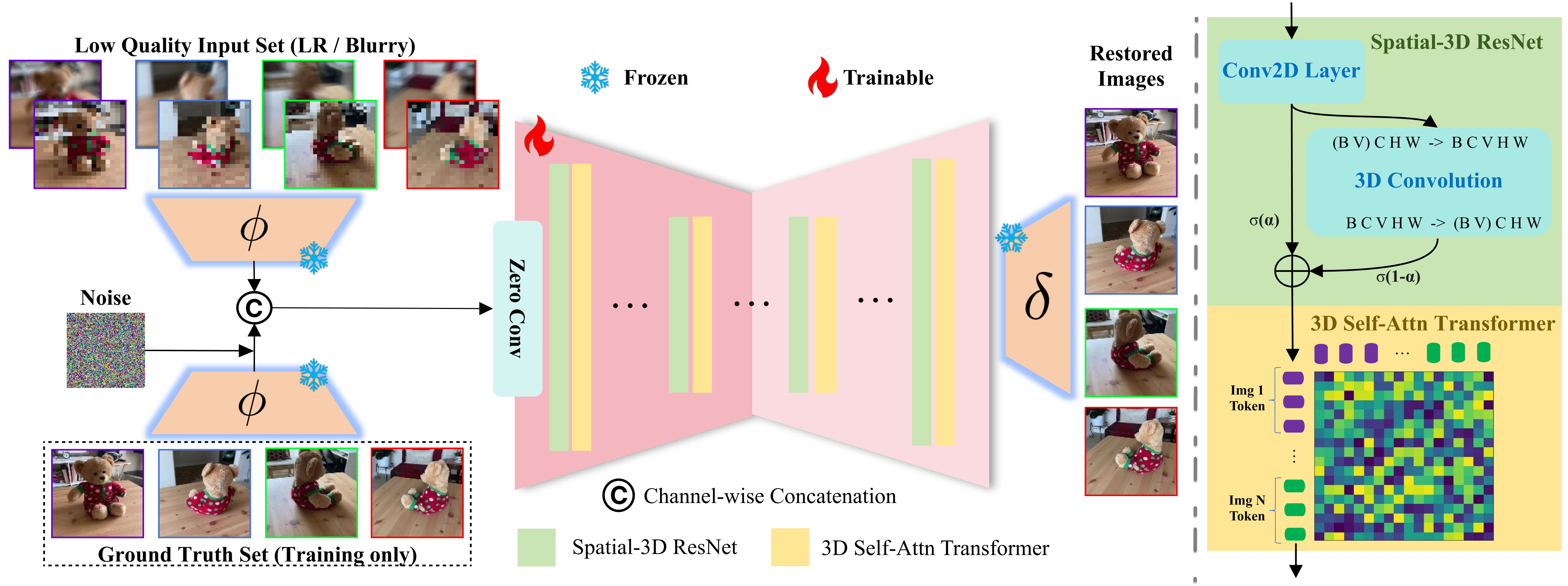}  
    \vspace{-0.65cm}
    \caption{
    {\bf Method Overview.} Our model is a UNet-based latent diffusion model that takes as input a degraded sparse image set and outputs a set of restored consistent images. The core of our approach is a diffusion model that jointly denoises image latents across multiple views by relying on our spatial-3D ResNet and 3D Self-Attn transformer. We use pre-trained encoder-decoders from SD2.1~\cite{rombach2022high} and train our denoising UNet on those encoded latents. We show view-consistent results for the task of \textbf{Deblurring} and \textbf{Super-Resolution}.
    }
    
    \label{fig:approach}
    \vspace{-0.1cm}  
\end{figure*}

In this section, we discuss our proposed approach, \ours, which can generate and restore sparse image sets from 3D scenes or objects conditioned on their degraded inputs.
Specifically, we showcase our approach to two challenging tasks: motion deblurring and super-resolution.
In the following sections, we discuss our problem setup in \S~\ref{subsec:problem_setup}, multi-view image restoration diffusion model in \S~\ref{subsec:diffusion}, and our proposed model architecture in \S~\ref{subsec:model}.

\subsection{Multi-View Generation Setup}
\label{subsec:problem_setup}
Our goal is to restore sparse view images from degraded inputs. 
Typical methods to solve this problem work either on a single image~\cite{wu2024one, potlapalli2024promptir, rombach2022high} or require a coherent video~\cite{zhou2024upscale, liang2024vrt} as input. 
Toward the goal of sparse view image restoration, we propose a generative model that can respect the geometrical priors to perform the task of image restoration from blurry images or recovering high-resolution image data from its low-resolution counterpart.
Any single-image generative model that performs independent inference on each image cannot ensure consistency over the generated image set.

To alleviate these issues, we formulate this as a generation problem to infer all the required views of the scene by modeling the joint distribution using a multi-view diffusion model. Our model takes a degraded image set as input and recovers a high-quality self-consistent image set.

\subsection{Multi-View Image Restoration Diffusion Model}
\label{subsec:diffusion}
Our model aims to restore images when conditioned on a corrupted image set $\icondSet$. 
The model's primary objective is to restore a self-consistent high-quality image set.
We do this by using a multi-view image restoration diffusion model that can jointly restore the image set. 
In training, we start from a dataset of non-blurry / high-resolution images $\irestSet$ and apply the forward diffusion process. Our input representation to the N-view diffusion model is $\xB = \{ {\irest_{1}, \dots, \irest_{N}}\}$. Specifically, $\xB^{0}$ corresponds to a clean signal at the start of the forward diffusion process at step $0$. 
 
Our generative model learns to reverse this noising process by predicting the added noise at every timestep of the diffusion process.

\subsection{Model Architecture}
\begin{figure*}[t]
    \centering
    \vspace{-1mm}
a    \includegraphics[width=1.0\linewidth]{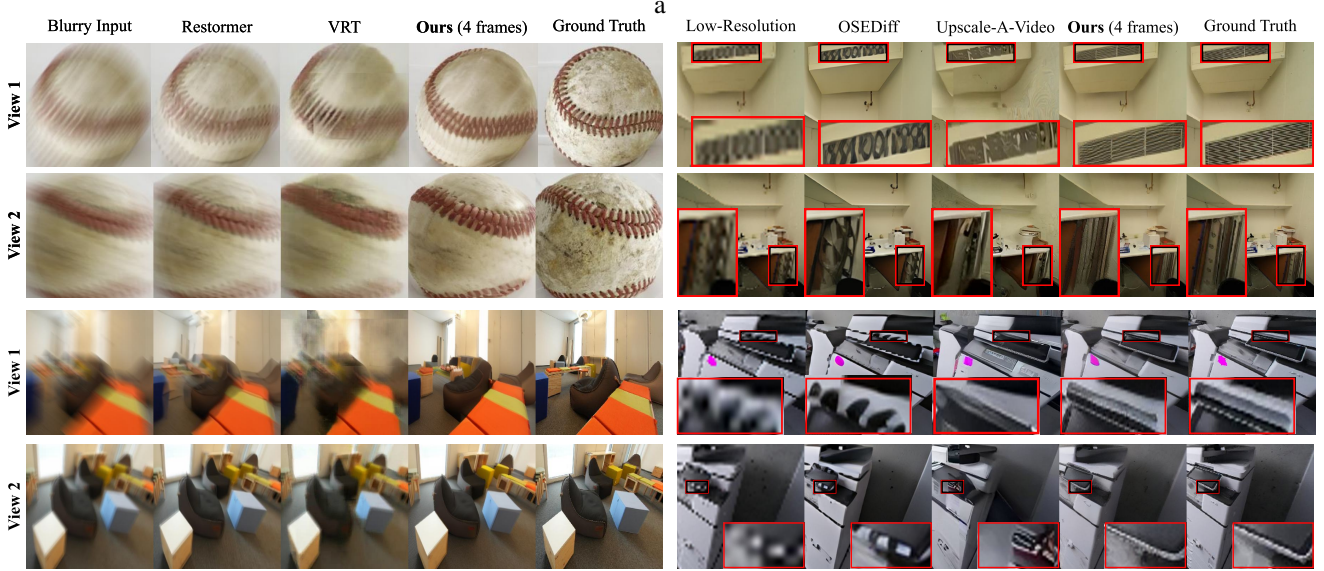}  
    \vspace{-0.7cm}
    \caption{
        \textbf{Qualitative Comparisons on Motion Deblurring and Super-Resolution.}
        The left side is the Motion-Deblurring and the right side is the Super-Resolution results.
        We refer readers to supplementary for more visualizations.
        \textbf{Zoom in for the best view.}
    }
    \label{fig:qualitative_visual}
    \vspace{-0.3cm}  
\end{figure*}

\label{subsec:model}

We build our joint image restoration diffusion model based on a latent diffusion model~\cite{rombach2022high}. Fig.~\ref{fig:approach} shows an overview of our method. 
The goal of our diffusion model is to denoise image latents such that a pre-trained decoder can convert them to high-quality images.
When simultaneously denoising multiple image latents, it is important that latent representation remain consistent across views. 
We enable this by allowing the diffusion model to attend to latents from other views.

\subsubsection{Latent Diffusion Models for Restoration}
A Latent Diffusion Model~\cite{rombach2022high} architecture, such as Stable Diffusion 2.1 (SD2.1), consists of two parts a) encoder and decoder and b) denoising UNet. 

During inference, our latent denoising UNet progressively refines pure noise into a clean latent representation, conditioned on the degraded image set, through several denoising steps. Once the latent representation is clean, a decoder translates it into the restored image set.

Similarly, during training, our diffusion model is optimized to learn the denoising process of the latent representation, conditioned on the degraded input sets.

\parnobf{Encoder-Decoder} We initialize our encoder with the weights from an SD2.1~\cite{rombach2022high} and are frozen when we train our model. 
The encoder, $\phi$, is used to map our images to latent vectors which are used to learn the UNet-Style diffusion model. 
The decoder, $\delta$, converts the denoised latents from the diffusion model to pixels in image space.

\parnobf{UNet} Our latent denoising UNet, $\sB_{\theta}(\cdot)$ with parameters $\theta$, considers as input a  noisy latent $\xB^{k}$,  encoded latents for  degraded images, $\icondSet$ and the diffusion noise $k$ to output an of estimate the noise, $\hat{\epsilonB}$, as shown in Eq. \eqref{eq:unet_denoise}:
\begin{equation}
    \label{eq:unet_denoise}
     \hat{\epsilonB} = \sB_{\theta}(\xB^{k}, \phi(\icondSet), k).
\end{equation}
We supervise the outputs of our UNet to match the noise added during the forward diffusion process using Eq.~\ref{eq:loss_function}:
\begin{equation}
    \label{eq:loss_function}    \mathcal{L} = \mathbb{E}_{ \mathbf{x}_0, \epsilon \sim \mathcal{N}(0,I),k \sim \mathcal{U}(T)} \left\| \mathbf{\epsilon} - \hat{\mathbf{\epsilon}} \right\|^2_2 .
\end{equation}

\subsubsection{Adapting Diffusion Model to multiple views}
Starting with the image diffusion model, we expand a new dimension to represent image sets from different views. Most of the time, batch and views are merged into one dimension and split when needed. 
We encode the degraded image set with the same VAE encoder as the target noisy latents. Noisy latents and the condition latents are integrated by channel-wise concatenation. Then, due to channel number change, we replace the original convolution with a new zero-initialized convolution layer~\cite{zhang2023adding} to adapt the new number of channels input.

\subsubsection{Spatial-3D Integrated UNet Architecture}
Typical latent diffusion models such as Stable Diffusion~\cite{rombach2022high} do not consider denoise multiple images simultaneously and often generate individual images that lead to inconsistent image generations between multi-views. 
Our solution to remove this inconsistency is to denoise all the views jointly by allowing the UNet to look at the intermediate latent representations from different views. 
As shown in the right side of Fig.~\ref{fig:approach}, our architecture is composed of a Spatial-3D ResNet and a 3D self-attention Transformer.

\parnobf{Spatial-3D ResNet}
Image diffusion model like SD2.1 only has a 2D convolution layer in the ResNet, which makes it hard to perform well in tasks that require 3D understanding. 
We want to enhance the 3D performance of the ResNet of the image diffusion model by introducing a 3D convolution layer to better capture geometry information as shown in Fig.~\ref{fig:approach}. 
Training from raw will be unstable and time-consuming, but we find that the video diffusion model that is performing the image-to-video task already has a pre-trained 3D convolution layer. Understanding temporal similarity in the time dimension and understanding 3D similarity in the spatial dimension share similarity. By the experiment, we empirically find that merging two different weights into one is feasible and can lead to better performance.  
Hence, we initialize the 2D Convolution layer with image Diffusion Model weight (SD2.1) and the 3D convolution layer with video Diffusion weight (Stable Video Diffusion~\cite{blattmann2023stable}). 
The output of each layer will be blended as follows: $O_{\text{ResNet}} =  \sigma(\alpha) \times O_{2D} + \sigma(1-\alpha) \times O_{3D}$,
where $\sigma$ is the sigmoid function and $\alpha$ is usually taken as 0.5 following SVD~\cite{blattmann2023stable}. $O_{2D}$ is the output of 2D ResNet and $O_{3D}$ is the output of 3D ResNet layer.

\begin{table*}[t]
\centering
\vspace{-0.1cm}
\caption{
     \textbf{Sparse-View Deblurring Comparisons.} We show quantitative results on three datasets and evaluate metrics as in \S~\ref{sec:metrics}. 
     VConsis. (visual consistency) is multiplied by 100. 
     Across all methods, \ours is the most visually consistent as compared to all the baselines. On other standard metrics, we also outperform baselines.
     \ours(1 Frame) is a single image version of our method that does not depend on multiple views. The best is {\bf highlighted}, and the second is \underline{underlined}. 
}
\vspace{-0.1cm}
\resizebox{0.8\textwidth}{!}{
\huge
\begin{tabular}{clccccccccc}
\cmidrule[\heavyrulewidth]{2-11}
 & \textbf{Dataset}
 & \multicolumn{3}{c}{\textbf{Scannet++}~\cite{yeshwanth2023scannet++}} & \multicolumn{3}{c}{\textbf{ETH3D}~\cite{schops2019bad}} & \multicolumn{3}{c}{\textbf{CO3D}~\cite{reizenstein2021common}} \\
 & Method & FID $\downarrow$ & LPIPS $\downarrow$ & VConsis.$\downarrow$ & FID $\downarrow$ & LPIPS $\downarrow$ & VConsis.$\downarrow$ & FID $\downarrow$ & LPIPS $\downarrow$ & VConsis.$\downarrow$ \\ \cmidrule{2-11}
 
& PromptIR~\cite{potlapalli2024promptir} &  81.28 & 0.248 & 7.81 & 150.34 & 0.537 & 10.72 & 166.3 & \underline{0.350} & 12.89  \\

& Restromer~\cite{zamir2022restormer}  & \underline{49.72} & \underline{0.232} & 6.52 & \underline{73.47} & \underline{0.330} & 9.59 & 144.8 & 0.426 & 11.00  \\

& VRT~\cite{liang2024vrt}  & 134.5 & 0.371 & 7.67 & 160.4 & 0.530 & 11.13 & 170.4 & 0.440 & 12.07 \\  

& \ours(1 Frame)  & 81.58 & 0.247 & \underline{6.45} & 109.6 & 0.350 & \underline{9.55} & \underline{137.9} & 0.357 & \underline{10.36} \\     

& \ours   & \textbf{40.09} & \textbf{0.160} & \textbf{5.75} & \textbf{67.31} & \textbf{0.263} & \textbf{9.43} & \textbf{85.67} & \textbf{0.275} & \textbf{9.30}
  
  \\ \cmidrule{2-11}

\end{tabular}
}
\label{tab:deblurring_table}
\vspace{-0.2cm}
\end{table*}

\parnobf{3D Self-Attention Transformer}
Furthermore, we adopt the 3D multi-view consistent self-attention Transformer as shown in Fig.~\ref{fig:approach}, similar to previous works~\cite{shi2023mvdream, gao2024cat3d}. 
Our self-attention layer considers image latents from all the $N$ views and performs self-attention. After ResNet, the model patches the latent to create $p$ patch token per view, and in total, we have $N \times p$ tokens that are part of the attention layer. Each token calculates the attention score to all $N \times p$ tokens as shown on the right side of Fig.~\ref{fig:approach}. 
This feature enables each token to be accessible to all conditional image sets given at the maximum level.
Further, by this design, our model, in the inference stage, is scalable to a variable number of input frames with stable performance even though we train with a handful of images per set.

However, attention communication between views leads to heavy computes. To resolve this, we only inject this module into the low-resolution layers (similar to CAT3D~\cite{gao2024cat3d}), where we also ignore the biggest resolution layer for the UNet Encoder and Decoder. Further, we completely take away the cross-attention module across all layers instead of putting empty string inputs as placeholders like previous works~\cite{ke2024repurposing}. We found that both of these changes help faster convergence, less training time, and better numerical results under the same number of training iterations.

\section{Experiment}

We evaluate \ours on the task of {\it sparse view} motion deblurring and super-resolution in a zero-shot setting \S~\ref{subsec:implementation}, and discuss our deblurring results in \S~\ref{subsec:motion_deblurring_result} and super-resolution results in \S~\ref{subsec:super_resolution_result}. We discuss implementation details in \S~\ref{subsec:implementation}, and now we start by discussing evaluation metrics in \S~\ref{sec:metrics}. Please refer to supplementary materials for the ablation study.

\subsection{Metrics}
\label{sec:metrics}
We use several metrics to evaluate the performance of all the methods on the task of motion deblurring and super-resolution. Since our approach is generative in nature we borrow metrics from the literature that are good at evaluating generative models such as {\it Frechet Inception Distance (FID)}~\cite{heusel2017gans}, {\it Learned Perceptual Image Patch Similarity (LPIPS)}~\cite{zhang2018unreasonable}. In addition to these, we evaluate our super-resolution results on PSNR and SSIM similar to methods used in this literature~\cite{wu2024one, zamir2022restormer, zhou2024upscale, yu2024scaling}. We also propose a new metric to evaluate view consistency between pairs of generated views.

\parnobf{Visual Consistency (VConsis)} This metric evaluates self-consistency within the generated view set by measuring the perceptual similarity of corresponding patches between the two views. Please refer to the supplementary for additional details.

\subsection{Implementation Details}
\label{subsec:implementation}
We implement our latent diffusion model based on pre-trained UNet and auto-encoder from SDv2.1 as part of diffusers.  We train our models on two synthetic datasets Hypersim~\cite{roberts2021hypersim} and TartanAir~\cite{wang2020tartanair} with image resolution resized to $640 \times 480$. 
We randomly select the order of image sets to maintain permutation invariance in training and inference. We utilize ground truth geometry and pose data to pre-compute the overlap region between each pair of frames. We provide additional details on model training, dataset, and view selection in the supplemental.

\subsection{Zero Shots Datasets For Evaluation}
As all our baselines are training on different datasets, we perform a zero-shot evaluation across a diverse set of real-world datasets to keep a fair comparison. {\it Zero-Shot} means rather than split the dataset into training-evaluation splits, we chose some dataset for training and some other out-of-domain for evaluation. Our evaluation datasets consist of indoor and outdoor scenes along with object-centric datasets. 

\parnobf{ScanNet++~\cite{yeshwanth2023scannet++}} This is a dataset of 460 indoor scanned readl 3D scenes. We evaluate all the methods on the scenes from the official test set of $50$ scenes, by sampling 200 image sets with four images per set. We highlight the zero-shot performance of our method trained on synthetic indoor data transfers well to real indoor scene data.

\parnobf{ETH3D~\cite{schops2019bad}} This dataset provides high-resolution images of indoor and outdoor scenes, with accurate depth maps captured via a LiDAR. We evaluate all the 454 samples from the ETH3D dataset.

\parnobf{CO3Dv2~\cite{reizenstein2021common}} This is an object-centric dataset containing object-level videos across 51 categories of household objects. This dataset has pose annotations from COLMAP~\cite{schonberger2016structure} provided for all video sequences.  We use the official single-sequence subsets for evaluation and randomly sample 200 image sets among them.

 \noindent For all these three datasets, please refer to the supplementary material for the detail of view-selection strategy

\subsection{Motion Deblurring (MD)}
\label{subsec:motion_deblurring_result}

\begin{table*}[t]
\centering
\vspace{-0.1cm}
\caption{
    \textbf{Sparse-View Super-Resolution Comparisons.} 
    VConsis. (visual consistency) is multiplied by 100. 
    \textbf{\ours(1 Frame)} is a single image process form like other image diffusion models. Ours, in the inference stage, can adapt an arbitrary number of views. 
    The best is \textbf{highlighted} and the second is \underline{underlined}.
}
\vspace{-0.05cm}
\resizebox{0.98\textwidth}{!}{%
\huge
\begin{tabular}{clcccccccccccc}
\cmidrule[\heavyrulewidth]{2-14}
 & \textbf{Dataset}
 & \multicolumn{4}{c}{\textbf{Scannet++}~\cite{yeshwanth2023scannet++}} & \multicolumn{4}{c}{\textbf{ETH3D}~\cite{schops2019bad}} & \multicolumn{4}{c}{\textbf{CO3D}~\cite{reizenstein2021common}} \\
 
 & Method & PSNR $\uparrow$ & SSIM $\uparrow$ & LPIPS $\downarrow$ & VConsis.$\downarrow$ & PSNR $\uparrow$ & SSIM $\uparrow$ & LPIPS $\downarrow$ & VConsis.$\downarrow$ & PSNR $\uparrow$ & SSIM $\uparrow$ & LPIPS $\downarrow$ & VConsis.$\downarrow$ \\ \cmidrule{2-14}

& SAM-DiffSR~\cite{wang2024sam} & 24.13 & 0.216 & 0.758 & 6.94 & 19.16 & 0.317 & 0.495 & 8.28 & 23.90 & 0.246 & 0.626& 10.96  \\

& OSEDiff~\cite{wu2024one} & 25.22 & \underline{0.808} & \underline{0.154} & \underline{5.51}  & 21.93 & \textbf{0.621} & 0.264 & 8.62 & \underline{24.50} & \textbf{0.658} & \underline{0.277} & \underline{8.86}  \\

& UAV~\cite{zhou2024upscale}  & 23.23 & 0.677 & 0.354 & 9.01 & 20.39 & 0.506 & 0.413 & 9.26 & 22.73 & 0.566 & 0.400 & 11.82\\  

& \textbf{\ours(1 Frame)}   & \underline{25.85} & 0.792 & 0.164 & 5.70 & \underline{21.97} & 0.593 & \underline{0.221} & \underline{7.90} & 24.35 & 0.641 & 0.301 & 9.07 \\     

& \textbf{\ours}   & \textbf{27.38} & \textbf{0.837} & \textbf{0.099} & \textbf{5.01} & \textbf{22.58} & \textbf{0.621} & \textbf{0.201} & \textbf{7.60} & \textbf{24.53} & \textbf{0.658} & \textbf{0.210} & \textbf{7.85}
  
  \\ \cmidrule{2-14}

\end{tabular}
}
\label{tab:sr_table}
\vspace{-0.2cm}
\end{table*}

Our motion deblurring (MD) method targets solving arbitrary motion paths in the wild. We target at restoring original images from the set of sparse view motion blur inputs.
Our paired training dataset is prepared by applying the motion-blurring kernel on the clean images from the training datasets based on the codebase~\footnote{https://github.com/LeviBorodenko/motionblur} inspired by DPS~\cite{chung2022diffusion}. 
The blur kernel size is chosen from a normal distribution with mean 85px and std 12.75px.

Further, the intensity of the blur is randomly sampled in the range $[0,1]$.
The blur kernel for each image in the set is sampled independently.
For evaluation, we randomly sample 4 images per view-set from the pre-computed image set and require all the methods to remove motion blurring from them.

\subsubsection{Baselines} 
We compare against several baselines that can restore motion blurring from single-image or videos.

\parnobf{Restormer~\cite{zamir2022restormer}} This is a single-image motion deblurring method that does not have the benefit of jointly deblurring across a sparse view image collection. Our method outperforms this on single image deblurring and sparse view deblurring across all metrics.

\parnobf{PromptIR~\cite{potlapalli2024promptir}} This is all-in-one restoration method trained to deblur. We use their pre-trained model and compare the performance of their method on sparse view deblurring.

\parnobf{VRT~\cite{liang2024vrt}} This is a video-oriented restoration Transformer model trained with deblurring capability. Our method outperforms this on sparse view deblurring across all metrics. Fair comparisons on video datasets are in the supplemental.

\subsubsection{Results}
We evaluate the performance of baselines and our method quantitatively and qualitatively on three datasets.

\parnobf{Qualitative} As shown in the left side of Fig.~\ref{fig:qualitative_visual}, our \ours show rich details and sharp deblurring restoration quality than other methods on both object level~\cite{reizenstein2021common} and scene level~\cite{yeshwanth2023scannet++}. 
This is thanks to our architectural design information from different views can be aggregated and constructive to help the other view in attention and residual layers. 

\parnobf{Quantitative} We assess our results using the metrics described in \S~\ref{subsec:implementation}, with outcomes summarized in Tab.~\ref{tab:deblurring_table}.

\noindent Our method outperforms all the methods on the three datasets on FID and LPIPS metrics. 
Restormer~\cite{zamir2022restormer}, a single-image method, lacks the ability to leverage multi-view information to enhance output quality. In contrast, our method uses 3D self-attention to achieve superior results.

We also outperform the video motion deblurring method, VRT, as they are incapable of handling sparse views well by motion estimation and require continuous and smooth video-captured views to fuse frame features.
In the supplement, we provide a direct comparison with video methods quantitatively on continuous videos.

\subsection{Super-Resolution (SR)}
\label{subsec:super_resolution_result}

We evaluate our method on the 4x sparse-view super-resolution task. 
Our paired training dataset is prepared by first downsampling the high-resolution image with a scale factor of 4 and then upsampling it back to the original resolution to introduce a low-resolution effect. The resize algorithm we use is bicubic.

\subsubsection{Baselines}
We compare against several diffusion model baselines that handle single-image and video super-resolution.

\parnobf{SAM-DIFFSR~\cite{wang2024sam}} This is an image diffusion model-based methodology, which targets classical super-resolution tasks like ours. 
Our method outperforms this on sparse view restoration on all the metrics.

\parnobf{OSEDiff~\cite{wu2024one}} This is a single-image diffusion model-based super-resolution model. 
OSEDiff reaches SOTA performance on the 2D image super-resolution.

\parnobf{Upscale-A-Video~\cite{zhou2024upscale}} This is a video diffusion model-based super-resolution method. We input it with a sequential video frame (in total 4 for each scenario) but a larger motion gap compared to ordinary videos.  We also provide a fair comparison between Upscale-A-Video and our method on video datasets in the supplement.

\subsubsection{Results}
We evaluate the performance of baselines and our method quantitatively and qualitatively on three datasets.
\parnobf{Qualitative} As the {\color{red}red} rectangular box of Fig.~\ref{fig:qualitative_visual} shows, 

for shadow artifacts and severe aliasing effects (see the second example), Upscale-A-Video~\cite{zhou2024upscale} fails and OSEDiff~\cite{wu2024one} restoration is unfaithful and inconsistent for different views compared to ours. 
Moreover, our model can restore images with better geometry alignment as shown in the first example.

\parnobf{Quantitative} We use PSNR, SSIM, LPIPS~\cite{zhang2018unreasonable} and our proposed visual consistency \S~\ref{sec:metrics} metrics. The result is shown in Tab.~\ref{tab:sr_table} to show that \ours outperforms all the existing baselines on all datasets.
Compared to the image diffusion method, OSEDiff\cite{wu2024one}, our spatial-3D Resnet and 3D self-attention Transformer help each latent token to access information from all views which leads to successful multiview restoration.
Upscale-A-Video also employs a similar 3D ResNet architecture and cross-image attention mechanism; however, our model does not rely on conditioning with optical flow, which is challenging to obtain accurately in sparse views.

\begin{table}[t]
\centering
\label{tab:badgs_hydrant_table}

\caption{
    \textbf{Gaussian Splatting Reconstruction from Blurry Images.}
    Under serious motion blurring, COLMAP\cite{schonberger2016structure} makes it hard to solve correct camera pose and SfM points, which leads to bad results on both vanilla 3DGS and BAD-GS. 
    With the help of \ours, we can achieve restore better results and be robust to difficult motion blurring effects.
    \textbf{KS} refers to kernel size with normal distribution [$\mu$, $\sigma$]. \textbf{Inten.} refers to the blurring intensity range that is sampled uniformly. We randomly select 3 objects in CO3Dv2~\cite{reizenstein2021common} dataset and apply motion blur on clean images to conduct this experiment.
    }
\vspace{-0.25cm}
\resizebox{1.0\linewidth}{!}{%
\huge
\vspace{-0.05cm}
\begin{tabular}{clcccccc}
\cmidrule[\heavyrulewidth]{2-8}
 & \textbf{Difficulty}
 & \multicolumn{3}{c}{Medium} & \multicolumn{3}{c}{Hard}\\
 & \textbf{Params}
 & \multicolumn{3}{c}{KS:[30, 10.2], Inten:[0, 0.4]} & \multicolumn{3}{c}{KS:[45,14.85], Inten:[0, 0.5]}\\
 & Method & PSNR $\uparrow$ & SSIM $\uparrow$ & LPIPS $\downarrow$ & PSNR $\uparrow$ & SSIM $\uparrow$ & LPIPS $\downarrow$ \\ \cmidrule{2-8}
& 3DGS~\cite{kerbl20233d} & 13.47 & 0.610 & 0.607 & 10.42 & 0.421 & 0.801  \\
& BAD-GS~\cite{zhao2024bad}  & 10.06 & 0.537 & 0.687 & 8.061 & 0.385 & 0.948\\  
& Resotrmer~\cite{zamir2022restormer}+BAD-GS~\cite{zhao2024bad}  & 25.77 & 0.616 & 0.356 & 23.88 & 0.560 & 0.317\\  
& \textbf{\ours + BAD-NeRF~\cite{badnerf}} & 23.60 & 0.503 & 0.452 & 22.47 & 0.439 & 0.485  \\
& \textbf{\ours + BAD-GS}  & \textbf{26.11} & \textbf{0.661} & \textbf{0.250} & \textbf{25.33} & \textbf{0.644} & \textbf{0.277}
  
  \\ \cmidrule{2-8}

\end{tabular}
}
\vspace{-0.0cm}
\label{tab:badgs_hydrant_table}
\end{table}

\subsection{Down Stream Applications}
We show several downstream applications of sparse multiview deblurring and super-resolution. Multi-view sparse images are used for tasks of finding \textbf{correspondences}, \textbf{pose estimation}, and \textbf{3D reconstruction}. 
The performance of these methods is often dependent on the input image quality. If the underlying image quality is degraded, the output of these methods is poor. 
Blindly restoring these images with off-the-shelf single-image methods leads to sub-optimal results because restoration from these methods lacks view consistency. We show that, with \ours, we can consistently restore image quality and help alleviate the performance gap on tasks requiring multiview sparse views.

\subsubsection{Estimating Correspondences}

\begin{figure*}[t]
    \centering
    \includegraphics[width=0.99\linewidth]{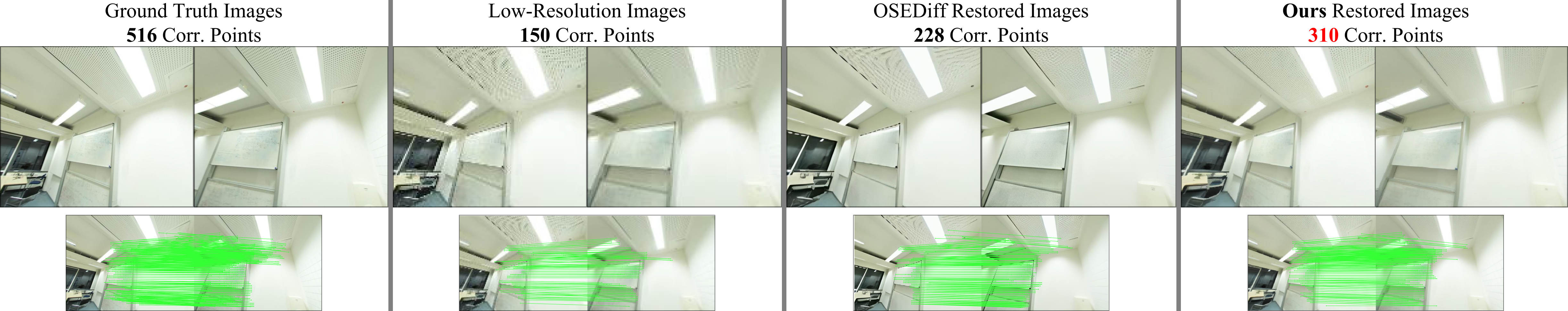}  
    \vspace{-0.25cm}
    \caption{
        \textbf{Feature Matching Downstream Application.} Low-Resolution and Single-image-based SR models like OSEDiff~\cite{wu2024one} find less correspondence points than our \ours that can restore multiview images consistently. \ours recovers the highest number of correspondences ($310$) as compared to other baseline restoration methods.
    }
    \label{fig:Lofter_application}
    \vspace{-0.2cm}  
\end{figure*}

\begin{table}[t]
\centering
\caption{
    \textbf{Feature Matching Correspondence.} 
    We compute the correspondences using off-the-shelf LoFTR~\cite{sun2021loftr} on the restored images with single-view methods and \ours. Both in SR and deblurring cases, our method can enable LoFTR to find more correspondences (\# of correspondences found).
}
\vspace{-0.2cm}
\resizebox{0.7\linewidth}{!}{%
\huge
\begin{tabular}{clcc}
\cmidrule[\heavyrulewidth]{2-4}

 & & Deblurring & SR \\
 \cmidrule{2-4}

& GT Images & 1979.5 & 1979.5 \\

\cmidrule[\heavyrulewidth]{2-4}

& Degraded Images & 640.1 & 1374.0 \\

& Single-View Methods~\cite{zamir2022restormer, wu2024one} & 1262.6 & 1478.0 \\

& \ours   & \textbf{1621.3} & \textbf{1494.2}
  
  \\ \cmidrule{2-4}

\end{tabular}
}
\label{tab:lofter_correspondence}
\vspace{-0.5cm}
\end{table}

Under severely degraded inputs it is hard for methods like LoFTR~\cite{sun2021loftr} to estimate correspondences. As shown in Fig.~\ref{fig:Lofter_application}, (a) we show qualitatively at low-resolution LoFTR struggles to find correct correspondences. Moreover, since single-view image restoration models lack multi-view consistency this does not solve the issue. \ours can not only restore the image quality but since it is view consistent can improve the correspondences for off-the-shelf methods like LoFTR~\cite{sun2021loftr}.
Quantitatively, as shown in Tab.~\ref{tab:lofter_correspondence}, we also see that LoFTR can find over $380$ additional correspondences as compared to the best image-based method on deblurring and on average 15 for super-resolution.

\subsubsection{3D Reconstruction on Degraded Images} 
\label{subsubsec:bad_gs_exp}

\begin{figure}[t]
    \centering
    \vspace{1mm}
    \includegraphics[width=1.0\linewidth]{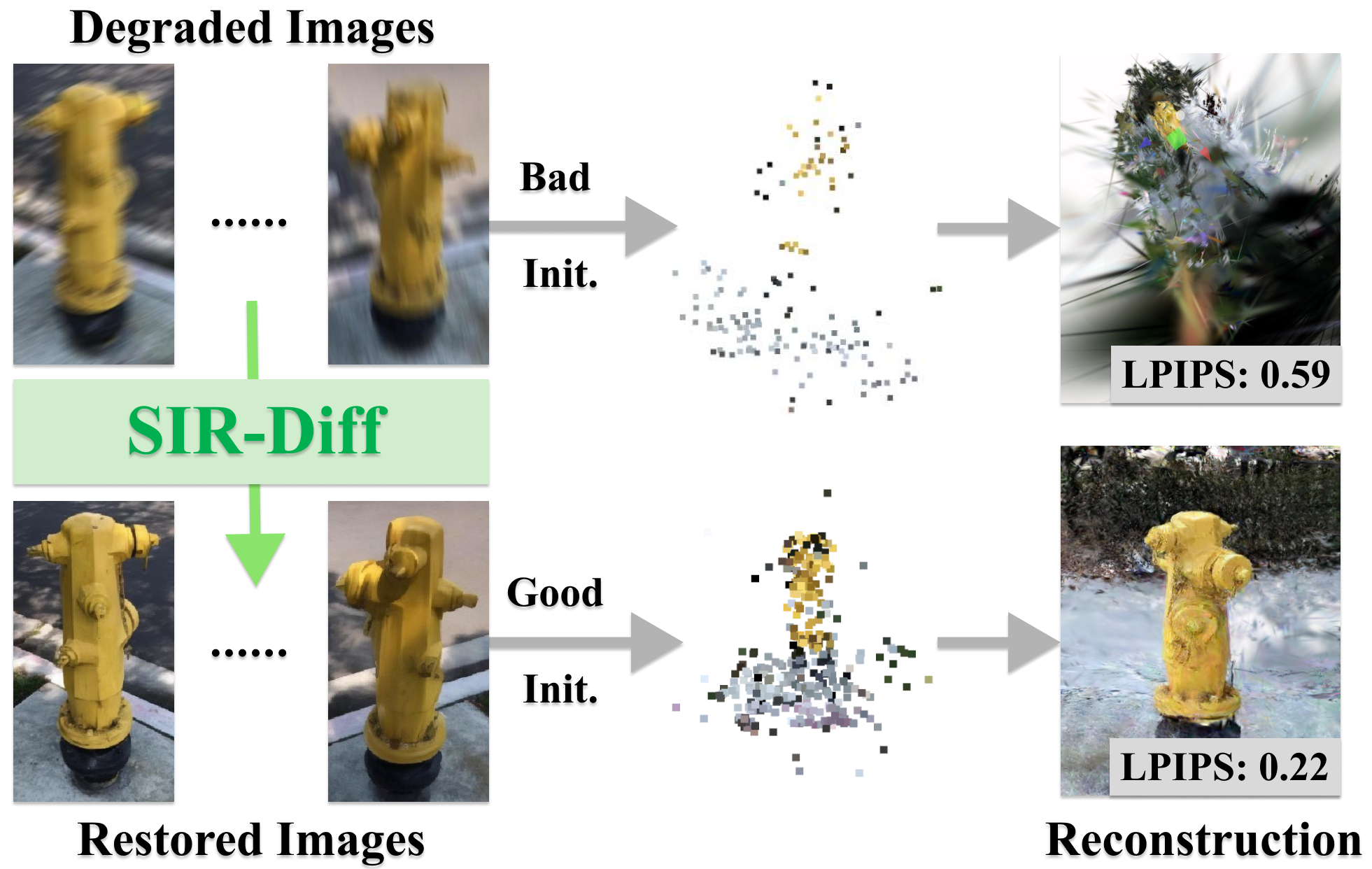}  

    \vspace{-0.4cm}
    \caption{
        \textbf{BAD-GS~\cite{zhao2024bad} with \ours.}  We show the effect of using \ours to help BAD-GS recover from catastrophic failure. BAD-GS (Top) output with a very high LPIPS$\downarrow$ of 0.59 \vs 0.22 BAD-GS output with image-restoration using \ours (Bottom) leads to a significantly higher-quality 3D reconstruction.
    }
    \label{fig:downStreamApplication}
    \vspace{-0.2cm}  
\end{figure}

Methods such as BAD-Gaussians~\cite{zhao2024bad} (BAD-GS) aim to perform 3D reconstruction from motion-blurry images, but they depend on running COLMAP~\cite{schonberger2016structure} on blurry images to get pose initialization and sparse 3D geometry. 
We show in Tab.~\ref{tab:badgs_hydrant_table} that with increasing blurring effect, BAD-GS performance drops significantly while using \ours with BAD-GS leads to a much steady performance. 
In Fig.~\ref{fig:downStreamApplication}, we show a hard case of motion blurring that leads to failure of BAD-GS~\cite{zhao2024bad} which can be solved if we restore the image using \ours first and then do BAD-GS. We refer to supplementary for details.

\subsubsection{Sparse-View 3D reconstruction}

\begin{figure}[t]
    \centering
    \vspace{1mm}
    \includegraphics[width=1.0\linewidth]{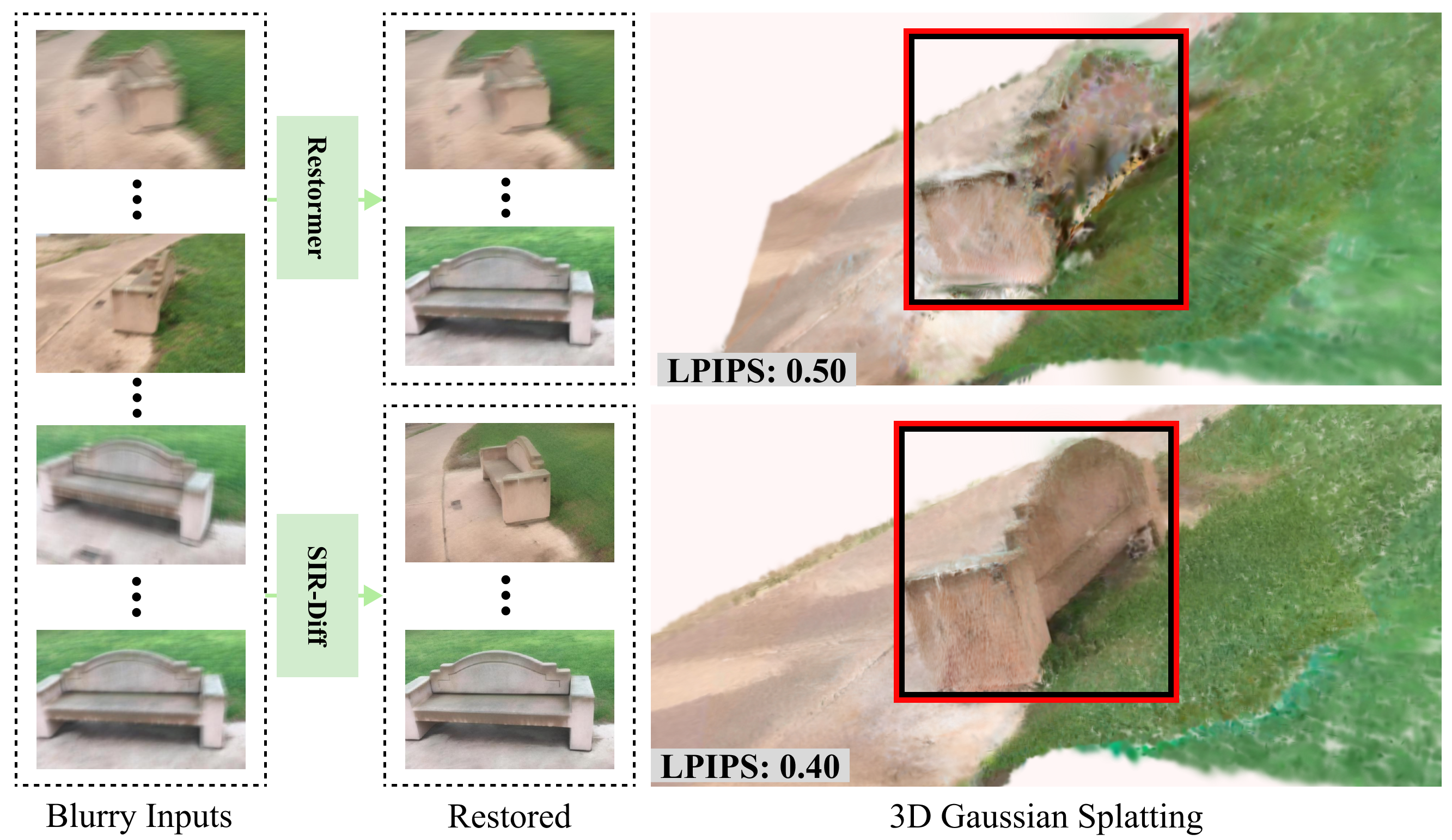}  
    \vspace{-0.5cm}
    \caption{
        \textbf{InstantSplat Downstream Application.} 3D Reconstruction is done by InstantSplat~\cite{fan2024instantsplat}. \ours that can restore multiview images consistently leads to better visual and numerical reconstruction results than other single-image methods, Restormer~\cite{zamir2022restormer}. 
    }
    \label{fig:InstantSplat_application}
    \vspace{-0.5cm}  
\end{figure}

InstantSplat~\cite{fan2024instantsplat} represents a leading approach for sparse view Gaussian Splattings (GS). However, reconstructing a scene under sparse viewpoints with low-quality observations remains challenging. 
In Fig~\ref{fig:InstantSplat_application}, we show an example of applying InstantSplat on an image restored using one of the baselines compared to restoration using \ours. 
In Tab.~\ref{tab:instantsplat_combined}, we report the average performance of all baselines on the selected 3 scenes from \textbf{Scannet++\cite{yeshwanth2023scannet++}} and \textbf{CO3Dv2\cite{reizenstein2021common}} and build a 3DGS  using InstantSplat~\cite{fan2024instantsplat}. We evaluate the GS with standard Novel View Synthesis metrics. Additional details can be found in the supplementary.

\begin{table}[t]
\centering

\label{tab:instantsplat_combined}
\caption{
    \textbf{Quantitative Results of Super-Resolution and Motion-Deblurring Reconstruction on InstantSplat \cite{fan2024instantsplat}.} 
    The result is an average of 3 scenes per dataset (individual results in supp). 
    Using \ours results in higher quality inputs to InstantSplat compared to OSEDiff~\cite{wu2024one} and Restormer~\cite{zamir2022restormer}.
}
\vspace{-0.2cm}
\resizebox{1.0\linewidth}{!}{
\vspace{-0.3cm}

\setlength{\tabcolsep}{2pt} 
\begin{tabular}{lccccccc}
\toprule
& & \multicolumn{3}{c}{\textbf{Scannet++~\cite{yeshwanth2023scannet++}}} & \multicolumn{3}{c}{\textbf{CO3D~\cite{reizenstein2021common}}} \\ 
& & PSNR\( \uparrow \)  & SSIM\( \uparrow \) & LPIPS\( \downarrow \)
& PSNR\( \uparrow \)  & SSIM\( \uparrow \) & LPIPS\( \downarrow \) \\
\hline
& GT Image & 31.86 & 0.940 & 0.096 & 23.94 & 0.757 & 0.243 \\
\hline
\multicolumn{8}{c}{\textit{Super-Resolution}} \\
\hline
& OSEDiff~\cite{wu2024one} & 24.53 & 0.799 & 0.274 & 20.25 & 0.549 & 0.379 \\
& \textbf{\ours} & \textbf{26.20} & \textbf{0.827} & \textbf{0.233} & \textbf{21.78} & \textbf{0.578} & \textbf{0.357} \\
\hline
\multicolumn{8}{c}{\textit{Motion-Deblurring}} \\
\hline
& Restormer~\cite{zamir2022restormer} & 22.02 & 0.789 & 0.309 & 20.14 & 0.579 & 0.465 \\
& \textbf{\ours} & \textbf{26.53} & \textbf{0.842} & \textbf{0.231} & \textbf{21.60} & \textbf{0.614} & \textbf{0.384} \\
\hline
\end{tabular}
}
\label{tab:instantsplat_combined}
\end{table}

\subsection{Depth Prediction for Sparse Views}

\begin{table}[t]
\centering
\caption{
    \textbf{Quantitative comparisons on image and video depth models.}
    The best is \textbf{highlighted} and the second is \underline{underlined}.
    GConsis. refers to geometric consistency and the unit is meter.
}
\vspace{-0.3cm}
\resizebox{1,0\linewidth}{!}{%
\huge
\begin{tabular}{clccccccccc}
\cmidrule[\heavyrulewidth]{2-8}
 & \textbf{Dataset}
 & \multicolumn{3}{c}{\textbf{Scannet++}~\cite{yeshwanth2023scannet++}} & \multicolumn{3}{c}{\textbf{ETH3D}~\cite{schops2019bad}} \\
 
 & Method & AbsRel $\downarrow$ & $\delta$1 $\uparrow$ & GConsis.$\downarrow$ & AbsRel $\downarrow$ & $\delta$1 $\uparrow$ & GConsis.$\downarrow$ \\ \cmidrule{2-8}

& DepthAnyV2~\cite{yang2024depth} & \underline{9.270} & \underline{93.35} & \underline{0.177} & 13.20 & 86.52 & 1.932 \\

& DepthCrafter~\cite{hu2024depthcrafter} & 21.65 & 89.96 & 0.751 & \textbf{6.662} & \textbf{95.05} & 1.701  \\

& Marigold~\cite{ke2024repurposing}  & 11.10 & 89.92 & 0.162 & 7.581 & 94.46 & \underline{0.410} \\   

& \textbf{\ours}   & \textbf{9.011} & \textbf{93.54} & \textbf{0.129} & \underline{6.663} & \textbf{95.05} & \textbf{0.339}
  
  \\ \cmidrule{2-8}

\end{tabular}
}
\label{tab:depth_table}
\vspace{-0.4cm}
\end{table}

We also apply our model architecture to the task of improving sparse multiview depth estimation. 
The only thing we edit is changing the target latent to depth image sets and the conditional latent to RGB image sets.
Multiview depth estimation is inherently a hard problem as it requires ensuring geometric consistency across multiple views when predicting the 3D geometry of the scene. Following Marigold ~\cite{ke2024repurposing}, we use the ground truth depth label provided by the dataset for supervision signal during training. 

\parnobf{Metrics} We evaluate the performance of methods on three metrics AbsRel and $\delta_{1}$ as defined by Silberman \etal~\cite{silberman2012indoor}.
We additionally propose a new metric to evaluate the consistency of depth prediction in sparse views. Specifically, we measure for the same point in 3D what is difference in depth value predicted in view any view $i$ and view $j$. We compute this metric by evaluating all views in pairs while accounting for occlusion. In the case of MDE methods, we additionally align all the predicted depth maps to find the most optimum scale and bias to make them consistent. Similar to the view consistency metrics for image restoration, We evaluate \textbf{Geometric Consistency} (GConsis.) by calculating the distances between corresponding points between two different views under the camera coordinate system. Additional details on this metric are in the supplemental.

\parnobf{Baselines} We evaluate against several monocular depth estimation (MDE) methods such as DepthAnythingV2~\cite{yang2024depth}, DepthCrafter~\cite{hu2024depthcrafter}, and Marigold~\cite{ke2024repurposing}. 
\parnobf{Results} 
As shown in Tab.~\ref{tab:depth_table}, our multi-view depth model is competitive with single image depth methods on the AbsRel, and $\delta_{1}$ metrics are sometimes better. Meanwhile, our model outperforms these methods on the geometric consistency metric that evaluates the 3D consistency when predicting depth jointly from sparse views.

\section{Conclusion \& Limitations}
\label{sec:conclusion}

We present \ours, a simple and effective approach for multi-view image restoration, addressing blurring and low-resolution degradation across multiple images simultaneously. \ours outperforms leading single-view methods in sparse-view restoration and enables 3D-consistent image restoration for 3D computer vision applications. Future work will explore efficient cross-image attention mechanisms, such as selective patch attention along epipolar lines, and develop explicit guarantees for 3D view consistency.

\section*{Acknowledgment}
This work partially used NCSA Delta GPU at the University of Illinois through allocation CIS230313 from the Advanced Cyberinfrastructure Coordination Ecosystem: Services \& Support (ACCESS) program, which is supported by U.S. National Science Foundation grants \#2138259, \#2138286, \#2138307, \#2137603, and \#2138296.

{
    \small
    \bibliographystyle{ieeenat_fullname}
    \bibliography{main}
}

\clearpage
\clearpage
\setcounter{page}{1}
\maketitlesupplementary

\section{Overview}

In Sec.~\ref{supp_qualitative}, we present comprehensive qualitative results for the tasks discussed in the main paper. 
Sec.~\ref{sec:supp_expriment} provides detailed descriptions of our training setup and explains the metrics used to evaluate our proposed method. 
In Sec.~\ref{sec:supp_video_compare}, we compare the performance of our method against video restoration methods using a standard video dataset. 
Further, Sec.~\ref{sec:supp_downstream} offers an in-depth explanation of our experiments on downstream applications, further demonstrating the versatility and effectiveness of our approach.
Finally, Sec.~\ref{sec:supp_ablation} offers the ablation study for our proposed method, and Sec.~\ref{sec:supp_additional_qualitative} provides additional qualitative results.

\section{Qualitative Visualization}\label{supp_qualitative}

\subsection{Real World Motion Debluring}
\label{sec:supp_main}

\begin{figure}[h]
    \centering
    \vspace{-2mm}
    \includegraphics[width=1.0\linewidth]{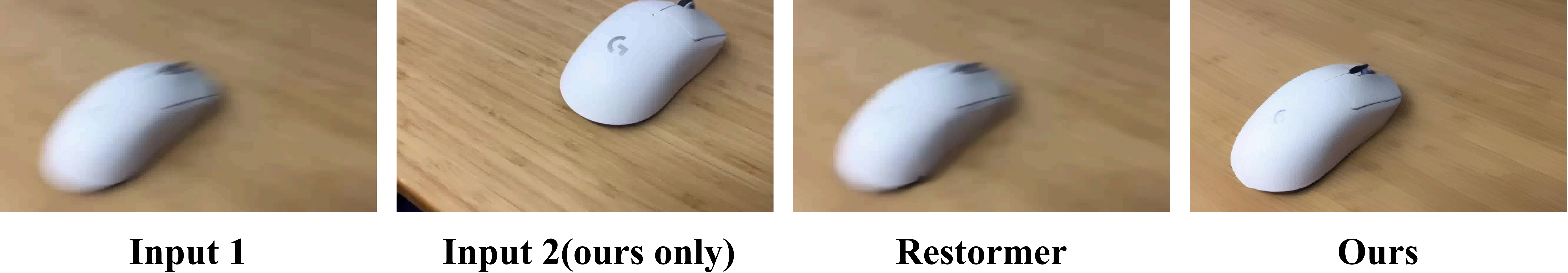}  
    \vspace{-8mm} 
    \caption{
        Qualitative comparison of \ours with Restormer on \textbf{in-the-wild real-world} images.
    }
    
    \label{fig:supp_real_world}
\end{figure}

\noindent In Fig.~\ref{fig:supp_real_world}, we show the performance of our model and baselines on deblurring the input on real images captured using a commodity smartphone. The deblurred output of our multi-view method is sharper than that of a single-view method Restormer \cite{zamir2022restormer}.

\subsection{Image Correspondences via LoFTR}
\label{sec:supp_main}

\begin{figure*}[h!]
    \centering
    \vspace{-0.3cm}
    \includegraphics[width=0.95\linewidth]{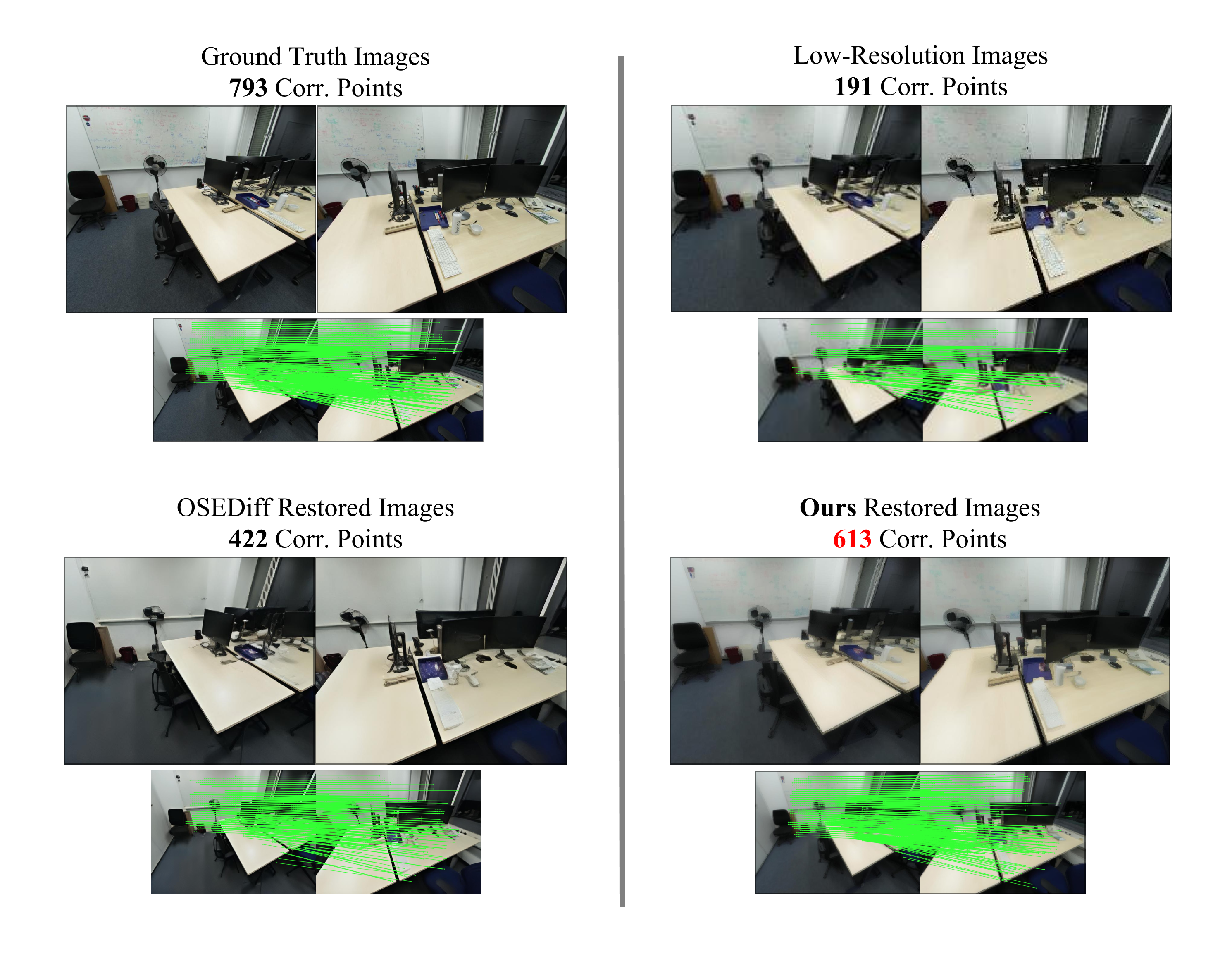}  
    \vspace{-0.7cm}
    \caption{
        \textbf{Correspondence matching of LoFTR~\cite{sun2021loftr} on low-resolution images.} We run a recent correspondence matching algorithm~\cite{sun2021loftr} on the restored images using our and a baseline method (OSEDiff~\cite{wu2024one}). Note that the algorithm fails to detect matches in blurred images. While Restormer processed images enable better matching, only about half of the matches are restored compared to the ground truth image pair. Restored images using our method produce significantly more matches, leveraging our multi-view denoising scheme.}
    \label{fig:supp_loftr_SR}
\end{figure*}

\begin{figure*}[h!]
    \centering
    \vspace{-0.2cm}
    \includegraphics[width=0.95\linewidth]{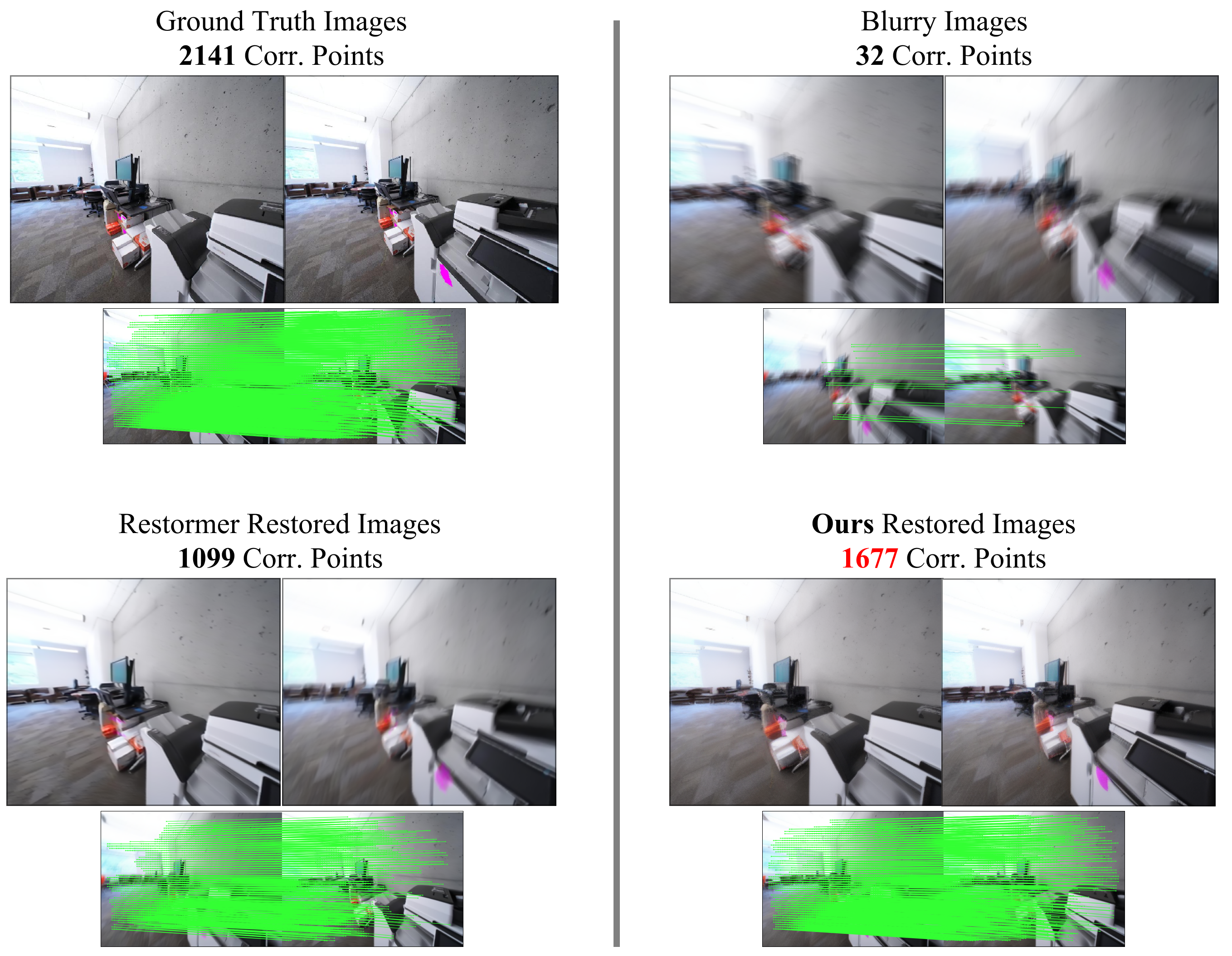}  
    \vspace{-0.25cm}
    \caption{
        \textbf{Correspondence matching of LoFTR~\cite{sun2021loftr} on blurry images.} We run a recent correspondence matching algorithm~\cite{sun2021loftr} on the restored images using our and a baseline method (Restormer~\cite{zamir2022restormer}). Note that the algorithm fails to detect matches in blurry images. While Restormer processed images enable better matching, only about half of the matches are restored compared to the ground truth image pair. Restored images using our method produce significantly more matches, leveraging our multi-view denoising scheme.
    }
    \label{fig:supp_loftr_deblur}
    \vspace{-0.4cm}  
\end{figure*}

In Fig.~\ref{fig:supp_loftr_SR} and Fig.~\ref{fig:supp_loftr_deblur}, we provide additional examples demonstrating how our model enhances the ability to LoFTR~\cite{sun2021loftr} model by identifying more corresponding points in low-resolution and motion-blurry sparse image sets.

\subsection{Gaussian Splatting on Motion Blurring Images}
\label{sec:supp_main}

\begin{figure*}[h!]
    \centering
    \vspace{-0.4cm}
    \includegraphics[width=0.99\linewidth]{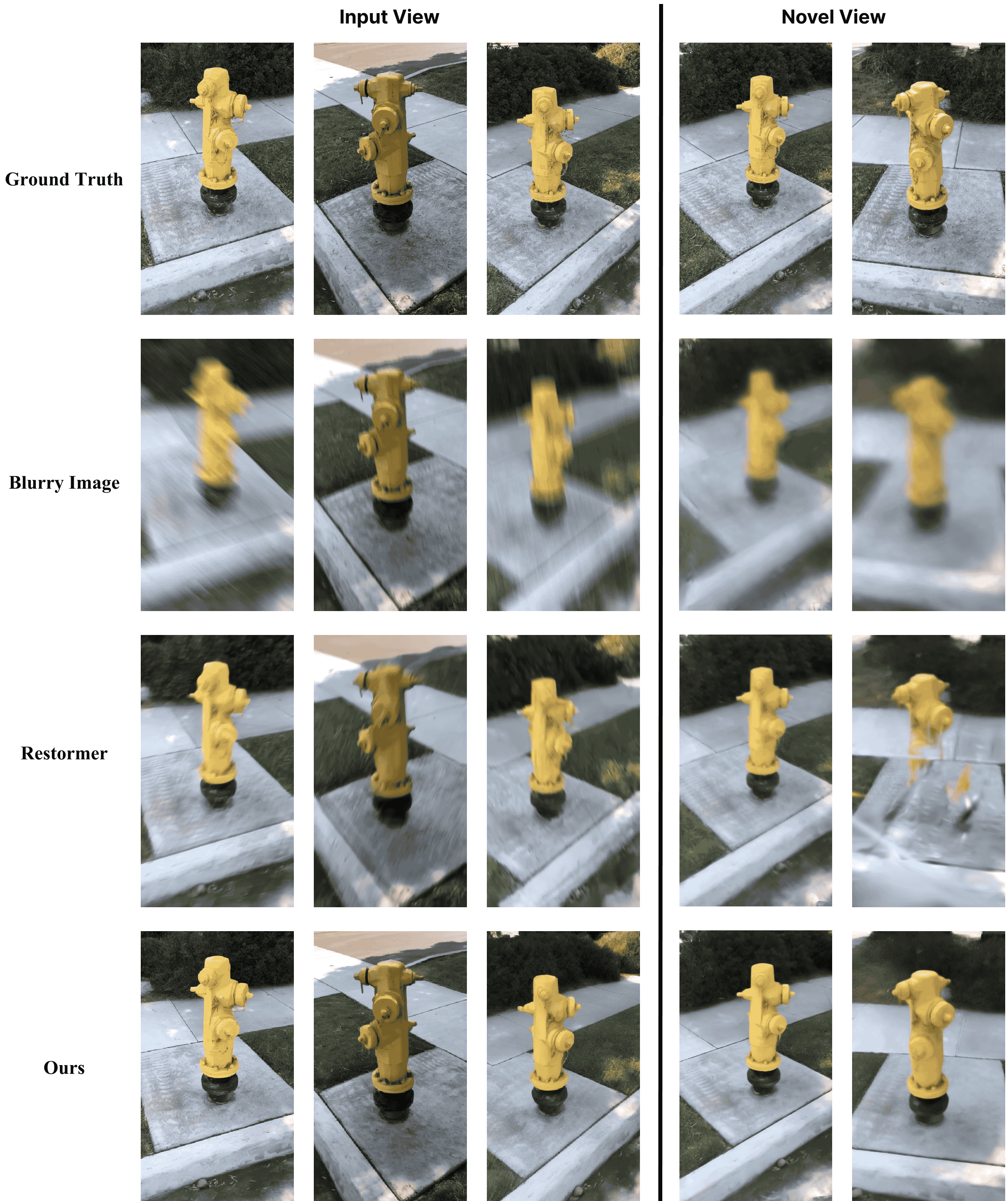}  
    \vspace{-0.25cm}
    \caption{\textbf{Gaussian Splatting Reconstruction Comparisons.} When we use 102 blurry images as inputs for GS, the rendered novel views exhibit strong artifacts (2nd row). The quality improves when using Restormer~\cite{zamir2022restormer} (3rd row) to deblur individual blurry images, but the rendering still includes artifacts compared to the ground truths (1st row). Our multi-view method (4th row) simultaneously deblurs all of the 102 images to produce consistent restorations, leading to higher-quality novel-view predictions.
    }
    \label{fig:supp_BAD_GS_training}
    \vspace{-0.4cm}  
\end{figure*}

Given a set of $N$ degraded input views in Fig~\ref{fig:supp_BAD_GS_training}, we use \ours and the baseline method to restore the input image views. Using the restored image views, we run BAD-GS, a state-of-the-art Gaussian Splatting reconstruction method from blurry inputs. We show the output-rendered images from a novel view. We show that \ours can restore the blurry images consistently which leads to better rendering of the Gaussian Splatting output from a novel view and faster converge speed during training.

\subsection{Sparse View Reconstruction from Degraded Images}
\begin{figure*}[t]
    \centering
    \vspace{-0.2cm}
    \includegraphics[width=0.9\linewidth]{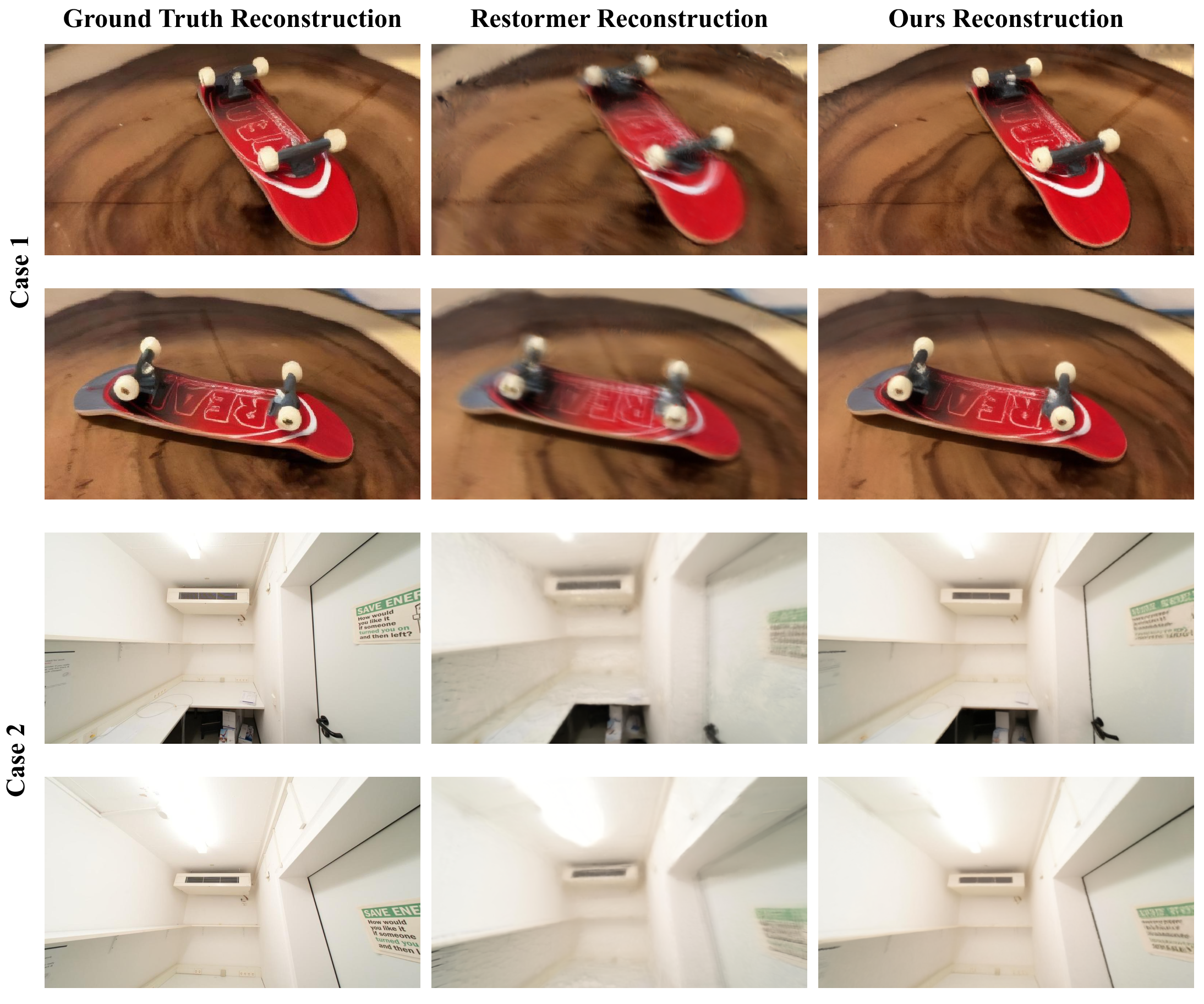}  
    \vspace{-0.25cm}
    \caption{
        \textbf{Sparse-View GS Reconstruction Results Using InstantSplat~\cite{fan2024instantsplat} on Motion Deblurring}.
        We jointly deblur 9 input blurry images using \ours (3rd column), leading to sharp reconstruction qualities compared to using Restomer restored images (2nd column). 1st column shows reference reconstruction using GT images.
    }
    \label{fig:supp_instantsplat_deblur}
    \vspace{-0.4cm}  
\end{figure*}

\begin{figure*}[t]
    \centering
    \vspace{-0.2cm}
    \includegraphics[width=0.9\linewidth]{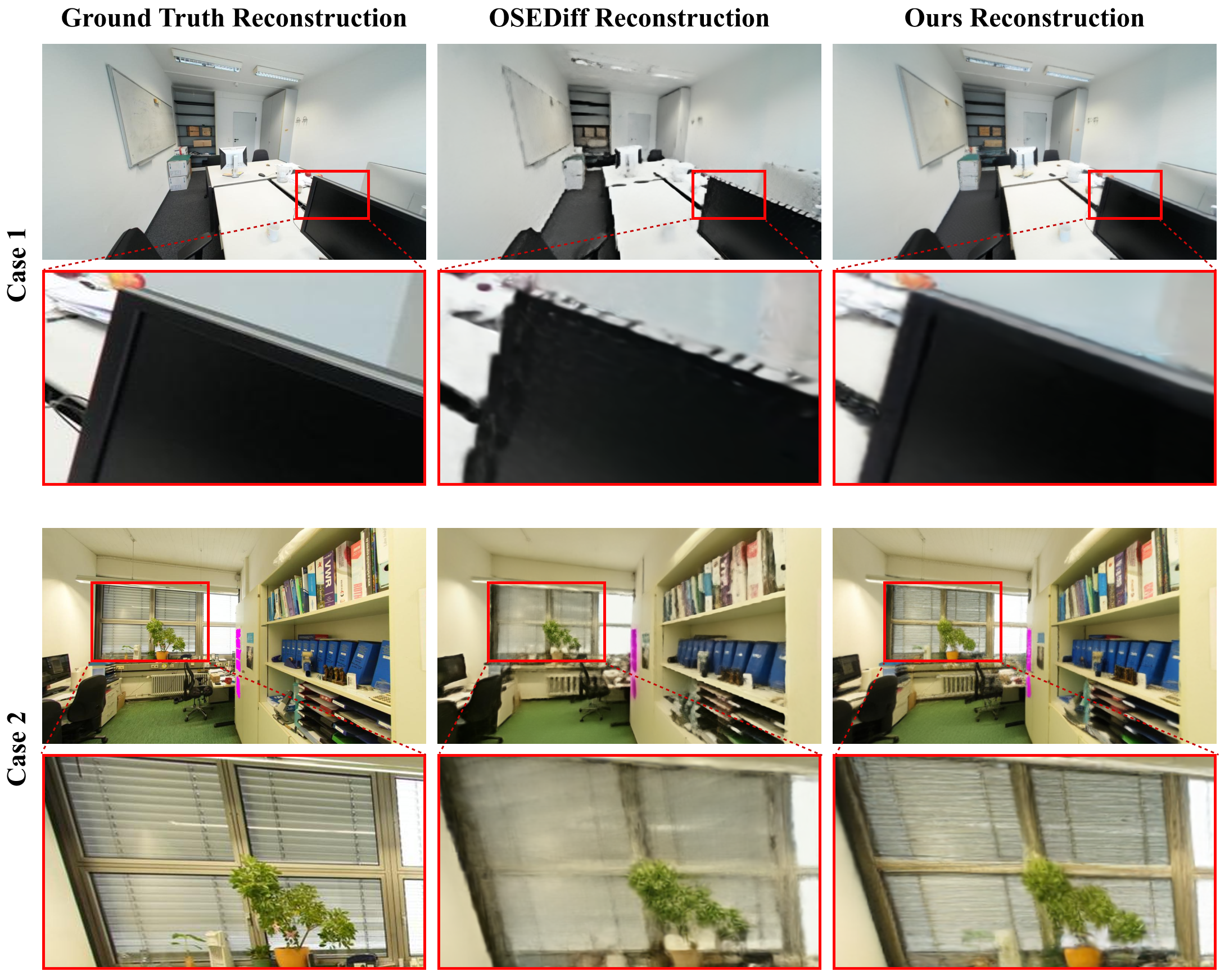}  
    \vspace{-0.25cm}
    \caption{
        \textbf{InstantSplat~\cite{fan2024instantsplat} Reconstruction Results for Super-Resolution.} We ran InstantSplat to obtain 3DGS on 9 images that were restored to be high-resolution, using OSEDiff (2nd column) and \ours (3rd column). As can be seen from the inset zoom images, our multi-view super-resolution leads to more 3D-consistent restoration, leading to sharper 3D reconstruction results closer to the ground truth reconstruction renderings (1st column).
    }
    \label{fig:supp_instantsplat_sr}
    \vspace{-0.2cm}  
\end{figure*}

In Fig.~\ref{fig:supp_instantsplat_deblur} and Fig.~\ref{fig:supp_instantsplat_sr}, we show the outputs reconstructing a sparse view 3D Gaussian Splatting method, InstantSplat~\cite{fan2024instantsplat}. 
We show that as compared to the baselines which do single view restoration for deblurring and super-resolution, \ours outshines them with much sharper and crisper results by enabling better sparse view reconstruction for InstantSplat~\cite{fan2024instantsplat}.
\label{sec:supp_main}

\section{Experiment Details}
\label{sec:supp_expriment}
In this section, we provide concrete details of our experiment, which is composed of metrics explanation in Sec.~\ref{subsec:supp_consis_metric} and training details in Sec.~\ref{subsec:supp_train_detail}.

\subsection{Metrics}
\label{subsec:supp_consis_metric}
To evaluate the self-consistency of the method, we propose the \textbf{Vision Consistency}, the metric that evaluates the consistency in RGB space, and the \textbf{Geometry Consistency} that evaluates the consistency of the 3D geometry.

\parnobf{Visual Consistency} We wish to evaluate the visual consistency between the restored images. The goal of this metric is to measure if two views that are geometrically consistent are also visually consistent.

Given two restored images, their ground truth depth maps and camera poses. We compute corresponding points between the images by establishing 3D correspondence using their 3D geometry. Naively measuring the difference in pixel values between corresponding pixels on the restored images is insufficient due to lighting and specularities.

To address these issues, we propose a method to evaluate the visual consistency of an image set. We compute ground-truth correspondences between two images, using the ground-truth geometry (depth) and pose, masking out all points within occluded regions. Each image is then divided into patches of size $30 \times 30$. Patches containing fewer than 300 corresponding points are discarded to ensure reliable evaluation. Given that the ground-truth correspondences within a patch remain sparse, we leverage these points to solve for a 2D affine transformation matrix, which is subsequently used to warp the patch from the source view to the target view, then the perceptual loss (LPIPS~\cite{zhang2018unreasonable}) is computed between the warped ground-truth image patch and the target ground-truth image patch. Ground truth patches that really look like each other have a perceptual loss of less than $0.1$. We use these patches for evaluation.  This process is repeated across several patches and average patch-wise perceptual loss is reported as the final measure of consistency between the two images. The lower the value of the metric more consistent are the images.

\parnobf{Gemoetry Consistency}
We evaluate the geometric consistency between two generated depth maps by evaluating the consistency of depth estimation for our results in \S \textcolor{red}{4.7} (main paper). We consider corresponding points between two images by using GT depth. If the depth discrepancy between two points is $\ge 0.1$ meter, we consider that the points are not in correspondence but one point occludes the other point. This allows us to create an occlusion mask for the two views using the ground-truth geometry. 

We now use the occlusion mask to evaluate the consistency between the generated depth images using our downstream application \S \textcolor{red}{4.7.} of the main paper.
Now for all the non-occluded points, we warp the generated depth map from the source view to the target view and evaluate the L1 distance between the warped depth map and the target depth map. We report the average consistency error in Tab.~\textcolor{red}{6} (main paper). We compute this metric for pairs of views in our evaluating image set.

\begin{figure*}[t]
    \centering

    \includegraphics[width=1.0\linewidth]{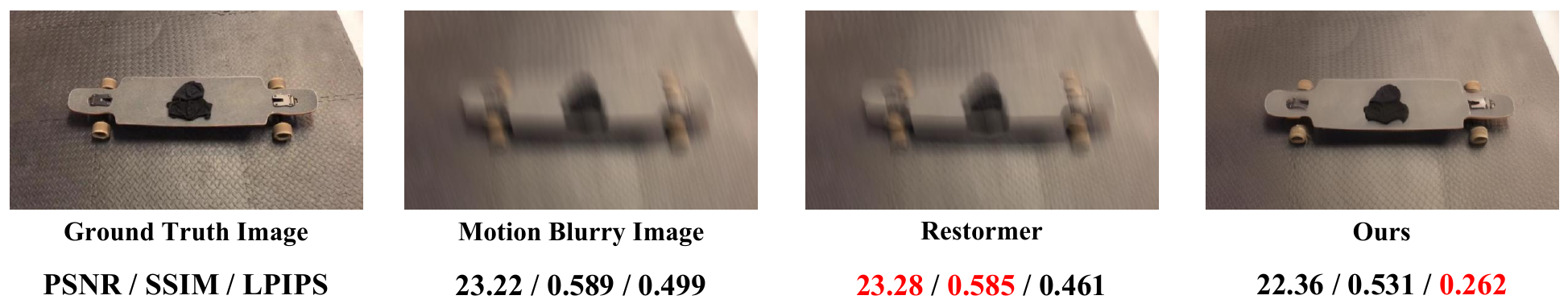}  
    \vspace{-0.5cm}
    \caption{
        \textbf{Ineffectiveness of PSNR and SSIM as Deblurring Metrics.}
        The best result is marked as {\color{red}red} for each metric. Ours show sharp and clear restored results but do not show an advantage on PNSR$\uparrow$ and SSIM$\uparrow$ than ~\cite{zamir2022restormer}, while LPIPS$\downarrow$ behaves as expected.
    }
    \label{fig:supp_psnrssim_figure}

\end{figure*}

\begin{figure*}[t]
    \centering

    \includegraphics[width=1.0\linewidth]{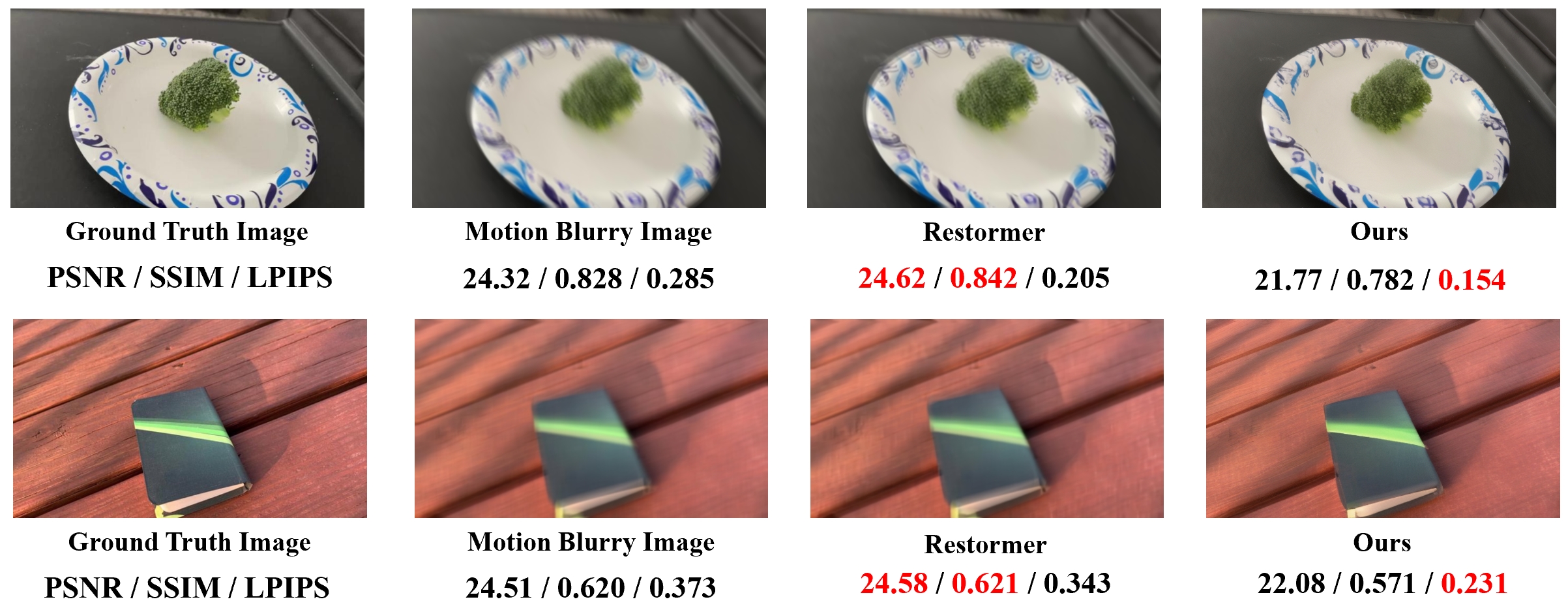}  
    \vspace{-0.25cm}
    \caption{
    \textbf{Ineffectiveness of PSNR and SSIM as Deblurring Metrics (Additional Examples).}
    The best result is marked as {\color{red}red} for each metric. Ours show sharp and clear restored results but do not show an advantage on PNSR$\uparrow$ and SSIM$\uparrow$, while LPIPS$\downarrow$ behaves as expected.
    }
    \label{fig:supp_more_psnrssim}

\end{figure*}

\parnobf{PSNR and SSIM for Deblurring} As mentioned in \S \textcolor{red}{4.4} of the main paper, for the evaluation of the Motion Deblurring task, we did not rely on traditional metrics such as PSNR and SSIM to assess the quality of image restoration. Previous study~\cite{zhang2018unreasonable} has shown that traditional metrics often do not strongly correlate with the perceived visual quality of images. We observed a similar issue in the Motion Deblurring task. As illustrated in Fig.~\ref{fig:supp_psnrssim_figure}, these metrics can fail to capture the visual differences between restored images effectively. Additional visualizations can be found in Fig.~\ref{fig:supp_more_psnrssim}. To address this limitation, we employed neural network-based perceptual metrics such as {\it Frechet Inception Distance (FID)}~\cite{heusel2017gans}, {\it Learned Perceptual Image Patch Similarity (LPIPS)}~\cite{zhang2018unreasonable} to evaluate the quality of image restoration in the Motion Deblurring task, providing a more robust and visually relevant assessment which can benefit the downstream task.

\subsection{Training Details}
\label{subsec:supp_train_detail}

We train our model on 2 synthetic datasets: Hypersim~\cite{roberts2021hypersim} and TartanAir~\cite{wang2020tartanair}. We use all the samples from Hypersim, around 50k high-quality RGB images. For TartanAir, we randomly sample around 10k image sets from the origin dataset. All the images we use for training are resized into 480 $\times$ 640 resolution. 
Training our model takes 30k iterations with a batch size of 8 for each GPU and each instance contains 4 views. We use 2 $\times$ NVIDIA A100 40GB or 2 $\times$ NVIDIA L40S GPU for training. We use the Adam optimizer with a learning rate of $3\cdot 10^{-5}$.

\parnobf{Image Set Selection Strategy}
We follow the same image set selection strategy in both training and testing datasets illustrated as follows: For each image in the dataset, we designate it as the anchor view. Next, we iterate through the remaining images and calculate the overlap region between the selected image and the anchor image. If the overlap ratio falls within the range of [0.6, 0.8] for training, the image is added to the precomputed image list corresponding to the anchor image. This selection process continues until the list contains 8 images per anchor. This selection algorithm is executed prior to training. During training, we randomly sample 4 images per instance from the precomputed image list as the image set. For testing, we randomly select 4 frames from 20 frames which near by the reference frames. This view selection strategy helps the model implicitly learn geometric relationships across the image sets, leading to improved performance. For all the images selected in the images set, we random shuffle the order before training to enable this permutation invariance at inference.

\section{Comparison on Traditional Video Dataset}
\label{sec:supp_video_compare}
As mentioned in \S \textcolor{red}{4.4} and \S \textcolor{red}{4.5} of the main paper, our main experiment shows the challenges faced by video-based image restoration methods when applied to image sets with large motion gaps and unordered inputs, as opposed to the smoothed and ordered structure of ordinary videos. Methods such as Upscale-A-Video~\cite{zhou2024upscale} and VRT~\cite{liang2024vrt} exhibit poor performance under these circumstances. 

To ensure a fair comparison with video-based methods, we conducted experiments on traditional video datasets. Since the ScanNet++~\cite{yeshwanth2023scannet++} dataset is also captured in a video format, it was included in our evaluation. Additionally, we incorporated two traditional video restoration datasets: youHQ~\cite{zhou2024upscale} and REDS~\cite{liu2022video}, which were used to train Upscale-A-Video~\cite{zhou2024upscale} and VRT~\cite{liang2024vrt}, while our method was evaluated in a \textbf{zero-shot} setting.

For the ScanNet++ dataset, we selected 24 consecutive frames from each of the 50 scenes in the evaluation set, resulting in a total of 50 videos for the evaluation. 
For the youHQ~\cite{zhou2024upscale} and REDS~\cite{liu2022video} datasets, we used their official evaluation splits and compared our results with the ones reported in their respective papers. Please refer to Tab.~\ref{tab:supp_video_sr} for results compared to the video-based super-resolution method and Tab.~\ref{tab:supp_video_deblur} for results compared to the video-based motion deblurring method.

\begin{table}[t]
\centering
\caption{
    \textbf{Comparison on Video Super-Resolution Dataset.} We present results on two video datasets: Scannet++~\cite{yeshwanth2023scannet++} and youHQ~\cite{zhou2024upscale}. Note that for the Scannet++~\cite{yeshwanth2023scannet++} dataset, both methods operate in a zero-shot setting. In contrast, for the youHQ~\cite{zhou2024upscale} dataset, Upscale-A-Video~\cite{zhou2024upscale} is evaluated in-domain, while our method remains in a zero-shot setting.}

\vspace{-0.2cm}
\resizebox{1.0\linewidth}{!}{%
\begin{tabular}{clcccccc}
\cmidrule[\heavyrulewidth]{2-8} &
& \multicolumn{3}{c}{Scannet++} & \multicolumn{3}{c}{youHQ} \\
& & PSNR $\uparrow$ & SSIM $\uparrow$ & LPIPS $\downarrow$ & PSNR $\uparrow$ & SSIM $\uparrow$ & LPIPS $\downarrow$ \\
\cmidrule{2-8}
& Upscale-A-Video~\cite{zhou2024upscale} & 26.41 & 0.8343 & 0.22 & 25.83 & 0.733 & 0.268 \\
& \ours  & 27.08 & 0.8456 & 0.1749 & 19.55 & 0.5249 & 0.4659
  
  \\ \cmidrule{2-8}

\end{tabular}
}
\vspace{-0.3cm}
\label{tab:supp_video_sr}
\end{table}

\begin{table}[t]
\centering
\caption{\textbf{Comparison on Video Motion-Deblurring Dataset.} We present results on two video datasets: Scannet++~\cite{yeshwanth2023scannet++} and REDS~\cite{liu2022video}. Note that for the Scannet++~\cite{yeshwanth2023scannet++}, both methods operate in a zero-shot setting. In contrast, for the REDS~\cite{liu2022video} dataset, VRT~\cite{liang2024vrt} is evaluated in-domain, while our method remains in a zero-shot setting. }
\vspace{-0.1cm}
\resizebox{1.0\linewidth}{!}{%

\begin{tabular}{clcccccc}
\cmidrule[\heavyrulewidth]{2-8} &

& \multicolumn{3}{c}{Scannet++} & \multicolumn{3}{c}{REDS} \\
& & PSNR $\uparrow$ & SSIM $\uparrow$ & LPIPS $\downarrow$ & PSNR $\uparrow$ & SSIM $\uparrow$ & LPIPS $\downarrow$ \\

\cmidrule{2-8}
& VRT~\cite{liang2024vrt} & 25.42 & 0.8470 & 0.2855 & 36.79 & 0.9648 & - \\

& \ours  & 26.72 & 0.8351 & 0.1840 & 19.60 & 0.4948 & 0.3465
  
  \\ \cmidrule{2-8}

\end{tabular}
}
\label{tab:supp_video_deblur}
\vspace{-0.3cm}
\end{table}


\section{Down Stream Applications}
\label{sec:supp_downstream}

\subsection{Estimating Correspondences}
We use the pre-trained LoFTR~\cite{sun2021loftr} model, trained on the {\it Indoor Dataset}, for evaluation. The same samples in Scannet++~\cite{yeshwanth2023scannet++} dataset from the main experiment are utilized. For each image set, we randomly sample two images to compute correspondences. 

First, we run the LoFTR~\cite{sun2021loftr} model on ground-truth high-quality RGB images to determine the number of correspondences that the model can find under ideal conditions. Then, we apply a single-view image restoration model and our proposed SIR-Diff to restore the degraded images. The LoFTR~\cite{sun2021loftr} model is subsequently run on the restored images to compute the number of correspondences, which serves as an indicator of the quality of the restored images. 
The results are presented in Tab.4 of the main paper.

\subsection{Gaussian Splatting on Reconstruction Motion Blurring Images}
\begin{figure*}[t]
    \centering
    \includegraphics[width=0.98\linewidth]{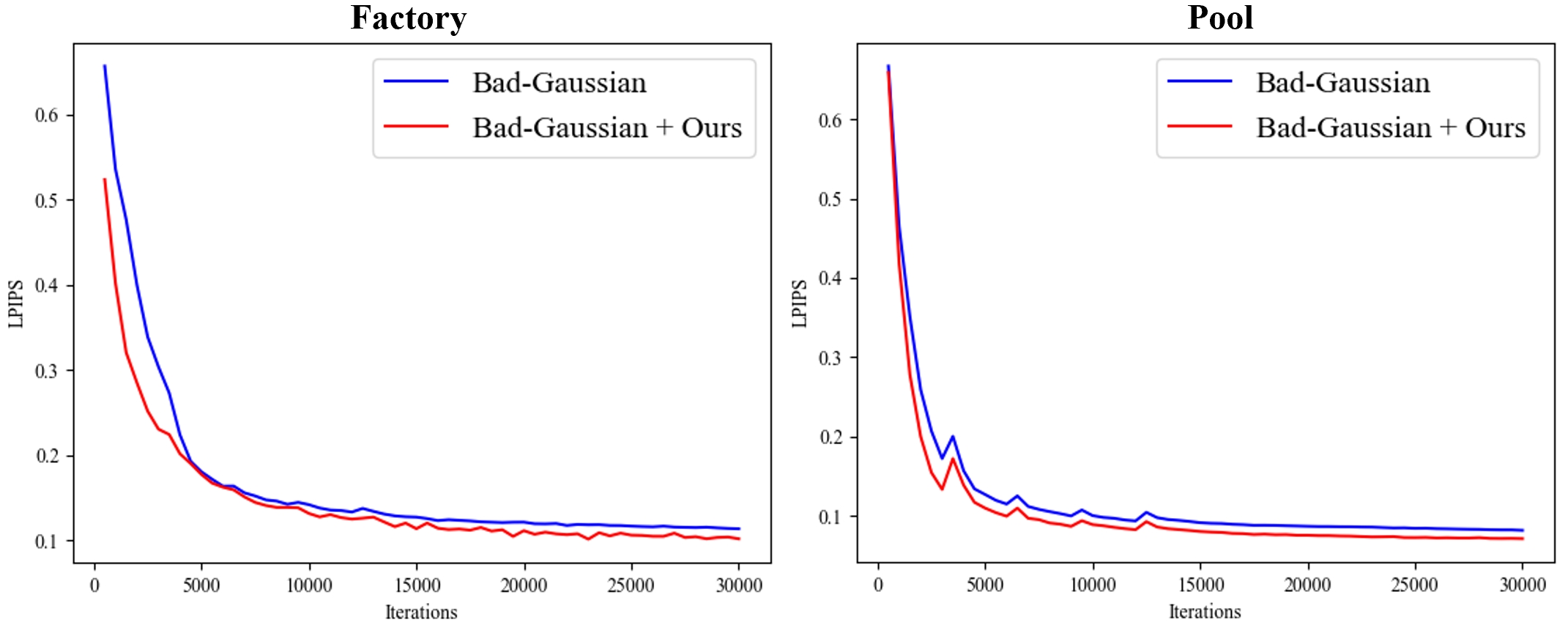}  
    \vspace{-0.3cm}
    \caption{
        {\bf Reconstruction Accuracy vs. Iterations.} 
        Applying our multi-view deblurring helps the BAD-Gaussians~\cite{zhao2024bad} algorithm to converge faster and improve reconstruction accuracy measured by LPIPS on the two scenes from ~\cite{ma2022deblur}. 
    }
    \label{fig:lpips_curve}
    \vspace{-0.1cm}
\end{figure*}

\begin{table*}[t]
\centering

\caption{
    \textbf{Per-scene result of Deblur-NeRF dataset}~\cite{ma2022deblur}. The best result is highlighted.
}
\vspace{-0.2cm}
\resizebox{1.0\linewidth}{!}{%
\huge

\begin{tabular}{clccc|ccc|ccc|ccc}
\cmidrule[\heavyrulewidth]{2-14}
 & \textbf{Difficulty}
 & \multicolumn{3}{c}{cozyroom} & \multicolumn{3}{c}{tanabata} & \multicolumn{3}{c}{Pool} & \multicolumn{3}{c}{Factory}\\
\cmidrule{2-14}
 & Method & PSNR $\uparrow$ & SSIM $\uparrow$ & LPIPS $\downarrow$ & PSNR $\uparrow$ & SSIM $\uparrow$ & LPIPS $\downarrow$ & PSNR $\uparrow$ & SSIM $\uparrow$ & LPIPS $\downarrow$ & PSNR $\uparrow$ & SSIM $\uparrow$ & LPIPS $\downarrow$ \\ 

& BAD-GS~\cite{zhao2024bad}  & \textbf{31.16} & \textbf{0.932} & 0.042 & \textbf{24.03} & \textbf{0.777} & 0.125 & \textbf{32.36} & \textbf{0.896} & 0.104 & \textbf{29.61} & \textbf{0.895} & 0.115\\  
& \ours + 3DGS~\cite{kerbl20233d} & 25.83 & 0.825 & 0.091 & 21.09 & 0.654 & 0.258 & 28.79 & 0.800 & 0.126 & 23.37 & 0.707 & 0.200 \\
& \textbf{\ours + BAD-GS}  & 30.38 & 0.931 & \textbf{0.041} & 21.96 & 0.698 & \textbf{0.125} & 31.97 & 0.891 & \textbf{0.082} & 28.37 & 0.834 & \textbf{0.104}
  
  \\ \cmidrule{2-14}

\end{tabular}
}
\label{tab:badgs_dbnerf_exp}
\end{table*}

\begin{table}[t]
\centering
\label{tab:supp_badgs_hydrant_table}
\caption{
    \textbf{Additional Gaussian Splatting Reconstruction Result from Blurry Images.} 
    The best is highlighted. 
    We additionally provide the results of the single-image motion deblurring model (Restormer~\cite{zamir2022restormer}) with BAD-GS~\cite{zhao2024bad}.
    }
\vspace{-0.25cm}
\resizebox{1.0\linewidth}{!}{%
\huge
\vspace{-0.15cm}
\begin{tabular}{clcccccc}
\cmidrule[\heavyrulewidth]{2-8}
 & \textbf{Difficulty}
 & \multicolumn{3}{c}{Medium} & \multicolumn{3}{c}{Hard}\\
 & \textbf{Params}
 & \multicolumn{3}{c}{KS:[30, 10.2], Inten:[0, 0.4]} & \multicolumn{3}{c}{KS:[45,14.85], Inten:[0, 0.5]}\\
 & Method & PSNR $\uparrow$ & SSIM $\uparrow$ & LPIPS $\downarrow$ & PSNR $\uparrow$ & SSIM $\uparrow$ & LPIPS $\downarrow$ \\ \cmidrule{2-8}
 & 3DGS~\cite{kerbl20233d} & 13.47 & 0.610 & 0.607 & 10.42 & 0.421 & 0.801  \\
& BAD-GS~\cite{zhao2024bad}  & 10.06 & 0.537 & 0.687 & 8.061 & 0.385 & 0.948\\ 
& Resotrmer~\cite{zamir2022restormer}+BAD-GS~\cite{zhao2024bad}  & 25.77 & 0.616 & 0.356 & 23.88 & 0.560 & 0.317\\  
& \textbf{\ours + BAD-GS~\cite{zhao2024bad}}  & \textbf{26.11} & \textbf{0.661} & \textbf{0.250} & \textbf{25.33} & \textbf{0.644} & \textbf{0.277}
  
  \\ \cmidrule{2-8}

\end{tabular}
}
\vspace{-0.0cm}
\label{tab:supp_badgs_hydrant_table}
\end{table}

\begin{table}[!htbp]
\centering
\caption{\textbf{Ablation Study.} The best is highlighted.}
\vspace{-0.3cm}
\resizebox{\columnwidth}{!}{%
\begin{tabular}{clcccccc}
\cmidrule[\heavyrulewidth]{2-6} &

Genre & PSNR $\uparrow$ & SSIM $\uparrow$ & LPIPS $\downarrow$ & Vconsis $\downarrow$
\\ \cmidrule{2-6}
& W/O 3D Convolution (4 Views) & 26.29 & 0.806 & 0.137 & 5.28 \\
& W/O SVD Init. (4 Views) & 26.28 & 0.813 & 0.192 & 6.03 \\
& 1 View & 25.85 & 0.792 & 0.164 & 5.70 \\
& 2 Views & 26.64 & 0.840 & 0.134 & 5.41 \\
& 4 Views & 27.38 & 0.837 & \textbf{0.099} & \textbf{5.01} \\
& 8 Views & \textbf{28.67} & \textbf{0.842} & 0.130 & 5.04 
\\ \cmidrule{2-6}
\end{tabular}
}
\label{tab:supp_ablation}
\end{table}

We follow the original experimental settings provided in the official implementation of BAD-Gaussians~\cite{zhao2024bad} (BAD-GS), which is based on NeRFStudio~\cite{nerfstudio}, and conduct experiments using the Deblur-NeRF dataset~\cite{ma2022deblur}. The per-scene result is presented in Tab.~\ref{tab:badgs_dbnerf_exp}.

Additionally, we observe that our method accelerates the convergence of the BAD-GS training process. To illustrate this, as shown in Fig.~\ref{fig:lpips_curve}, we plot the LPIPS~\cite{zhang2018unreasonable} loss curve during the training of BAD-GS on the {\it Factory} and {\it Pool} scenes. The results demonstrate that the incorporation of our model improves the convergence speed, underscoring its effectiveness in enhancing the training process.

As discussed in Sec.4.6.2 of the main paper, we also evaluate BAD-GS under high-intensity motion blurring conditions. To simulate motion blurring, we apply varying strengths of motion blur kernels to sharp images, use our proposed \ours to deblur the images, and then perform reconstruction using COLMAP. The results in Tab.3 of the main paper reveal that our method achieves robust and consistent performance across different intensity levels of motion blurring. 
While certain single-view methods also perform well, their lack of self-consistency in restored images results in inferior reconstruction performance compared to our approach. 
Detailed results can be found in Tab.~\ref{tab:supp_badgs_hydrant_table}.

For all experiments in this section, the reported results include not only the performance on \textbf{Novel View Synthesis} but also the performance on deblurring the \textbf{Training Views}. By considering both aspects, we provide a comprehensive evaluation of our method's effectiveness in addressing motion-blurring effects across the entire dataset.

\subsection{Sparse-View 3D reconstruction from Degraded Images}

We build our method based on the officially released code of InstantSplat~\cite{fan2024instantsplat}. We randomly select 3 scenes from Scannet++ dataset~\cite{yeshwanth2023scannet++} and 3 objects from CO3Dv2 dataset~\cite{reizenstein2021common}. 
Following the original setting of InstantSplat~\cite{fan2024instantsplat}, we first randomly sample 24 images as the training-evaluation set, then randomly sample 9 views from it as the training views and the rest views for evaluation of Novel View Synthesis. For the motion deblurring reconstruction task, we apply the same intensity of the blurring kernel on the ground-truth sharp images from training views in which the blurring kernel size is chosen from a normal distribution with mean 85px and standard deviation 12.75px, and the intensity of the blurring is randomly sampled in the range $[0,1]$. For the super-resolution reconstruction task, we downsample the original high-quality image with a ratio of 4. We use our \ours model and the best single-view image restoration mothed to restore the degraded image set and use it for reconstruction. All the experiment results are reported in \textbf{Novel View Synthesis} setting. We also report our per-scene result of super-resolution in Tab~\ref{tab:supp_quant_sr_instantsplat} and the per-scene result of Motion Deblurring in Tab~\ref{tab:supp_quant_deblur_instantsplat}.

\begin{table*}[t]
\centering
\caption{
    \textbf{Per-Scene Result for the Super-Resolution Reconsturction on InstantSplat~\cite{fan2024instantsplat}.} The best excluding the GT is highlighted.
}
\label{tab:supp_quant_sr_instantsplat}
\vspace{-0.2cm}
\resizebox{\linewidth}{!}{
\begin{tabular}{c | ccc | ccc | ccc}
\toprule
\multirow{2}{*}{Scannet++~\cite{yeshwanth2023scannet++}} & \multicolumn{3}{c|}{825d228aec} & \multicolumn{3}{c|}{acd95847c5} & \multicolumn{3}{c}{d755b3d9d8} \\
\cline{2-10}
& SSIM \( \uparrow \) & PSNR \( \uparrow \) & LPIPS \( \downarrow \) & SSIM \( \uparrow \) & PSNR \( \uparrow \) & LPIPS \( \downarrow \) & SSIM \( \uparrow \) & PSNR \( \uparrow \) & LPIPS \( \downarrow \) \\
\midrule
Ground Truth Image & 0.976 & 35.25 & 0.065 & 0.953 & 33.33 & 0.097 & 0.892 & 27.02 & 0.127 \\
\midrule

OSEDiff~\cite{wu2024one} & 0.917 & 26.78 & 0.217 & 0.812 & 26.08 & 0.243 & 0.669 & 20.75 & 0.364 \\
\textbf{Ours} & \textbf{0.923} & \textbf{28.41} & \textbf{0.163} & \textbf{0.876} & \textbf{27.39} & \textbf{0.204} & \textbf{0.684} & \textbf{22.82} & \textbf{0.334} \\
\midrule
\midrule
 
\multirow{2}{*}{CO3D~\cite{reizenstein2021common}}  & \multicolumn{3}{c|}{Bench} & \multicolumn{3}{c|}{Hydrant} & \multicolumn{3}{c}{Skateboard} \\
\cline{2-10}
& SSIM \( \uparrow \) & PSNR \( \uparrow \) & LPIPS \( \downarrow \) & SSIM \( \uparrow \) & PSNR \( \uparrow \) & LPIPS \( \downarrow \) & SSIM \( \uparrow \) & PSNR \( \uparrow \) & LPIPS \( \downarrow \) \\
\midrule
Ground Truth Image & 0.758 & 25.54 & 0.224 & 0.697 & 21.09 & 0.277 & 0.818 & 25.20 & 0.230 \\
\midrule

OSEDiff~\cite{wu2024one} & 0.486 & 20.82 & 0.393 & 0.466 & 17.98 & 0.417 & 0.695 & 21.96 & 0.329 \\
\textbf{Ours} & \textbf{0.502} & \textbf{22.70} & \textbf{0.371} & \textbf{0.508} & \textbf{19.49} & \textbf{0.396} & \textbf{0.725} & \textbf{23.15} & \textbf{0.305} \\
\bottomrule
\end{tabular}
}
\end{table*}

\begin{table*}[t]
\centering
\vspace{0.2cm}
\caption{
    \textbf{Per-Scene Result for the Motion Deblurring Reconsturction on InstantSplat~\cite{fan2024instantsplat}.} The best despite the GT is highlighted.
}
\label{tab:supp_quant_deblur_instantsplat}
\vspace{-0.2cm}
\resizebox{\linewidth}{!}{
\begin{tabular}{c | ccc | ccc | ccc}
\toprule
\multirow{2}{*}{Scannet++~\cite{yeshwanth2023scannet++}} & \multicolumn{3}{c|}{825d228aec} & \multicolumn{3}{c|}{acd95847c5} & \multicolumn{3}{c}{d755b3d9d8} \\
\cline{2-10}
& SSIM \( \uparrow \) & PSNR \( \uparrow \) & LPIPS \( \downarrow \) & SSIM \( \uparrow \) & PSNR \( \uparrow \) & LPIPS \( \downarrow \) & SSIM \( \uparrow \) & PSNR \( \uparrow \) & LPIPS \( \downarrow \) \\
\midrule
Ground Truth Image & 0.976 & 35.25 & 0.065 & 0.953 & 33.33 & 0.097 & 0.892 & 27.02 & 0.127 \\
\midrule

Restormer~\cite{zamir2022restormer} & 0.869 & 20.52 & 0.305 
& 0.830 & 24.95 & 0.253 
& 0.669 & 20.62 & 0.370 \\
\textbf{Ours} & \textbf{0.931} & \textbf{29.00} & \textbf{0.165} 
& \textbf{0.885} & \textbf{27.76} & \textbf{0.206} 
& \textbf{0.713} & \textbf{22.82} & \textbf{0.323} \\
\midrule
\midrule

\multirow{2}{*}{CO3D~\cite{reizenstein2021common}} & \multicolumn{3}{c|}{Bench} & \multicolumn{3}{c|}{Hydrant} & \multicolumn{3}{c}{Skateboard} \\
\cline{2-10}
& SSIM \( \uparrow \) & PSNR \( \uparrow \) & LPIPS \( \downarrow \) & SSIM \( \uparrow \) & PSNR \( \uparrow \) & LPIPS \( \downarrow \) & SSIM \( \uparrow \) & PSNR \( \uparrow \) & LPIPS \( \downarrow \) \\
\midrule
Ground Truth Image & 0.758 & 25.54 & 0.224 & 0.697 & 21.09 & 0.277 & 0.818 & 25.20 & 0.230 \\
\midrule

Restormer~\cite{zamir2022restormer}
& 0.516 & 20.62 & 0.490
& 0.400 & 11.00 & 0.696 
& 0.519 & 14.54 & 0.567 \\
\textbf{Ours} 
& \textbf{0.545} & \textbf{22.00} & \textbf{0.406}
& \textbf{0.498} & \textbf{18.34} & \textbf{0.476}
& \textbf{0.669} & \textbf{20.63} & \textbf{0.352} \\
\bottomrule
\end{tabular}
}
\vspace{-0.1cm}
\end{table*}
\section{Ablation Study}
\label{sec:supp_ablation}

To verify the effectiveness of each modality we propose, we do several ablation studies: 
(1) Does the 3D convolution layer in our spatial-3D ResNet help? 
(2) Does the SVD~\cite{blattmann2023stable} weight initialization on the 3D convolution layer help? 
(3) What is the performance variation with different numbers of views in the 3D self-attention Transformer during the inference?

We conduct an ablation study on the Scannet++ dataset~\cite{yeshwanth2023scannet++} with a super-resolution task. We used the same split and image set in the main experiment.
As shown in Tab.~\ref{tab:supp_ablation}, without a 3D convolution layer, the performance on 4 views restoration drops greatly on all metrics. This proves that only deploying the default 2D convolution layer from Latent Diffusion Models like SD2.1~\cite{rombach2022high} is limited in 3D tasks. 
However, the performance is still worse even if we have a 3D convolution layer, but don't initialize the weight with a reasonable pre-trained model.  From here, we also show that even though the scheduler that SVD uses (EDM~\cite{karras2022elucidating}) is different from the DDPM~\cite{ho2020denoising} that we use on the training, the stability of training and the performance improvement can still be maintained.
Furthermore, the results from 1 view to 8 views show that our model can handle different numbers of views, as the input and overall performance improvement are proportional to the number of views available.

\section{Additional Qualitative}
\label{sec:supp_additional_qualitative}

\begin{figure*}[t]
    \centering
    \vspace{-0.2cm}
    \includegraphics[height=0.98\textheight]{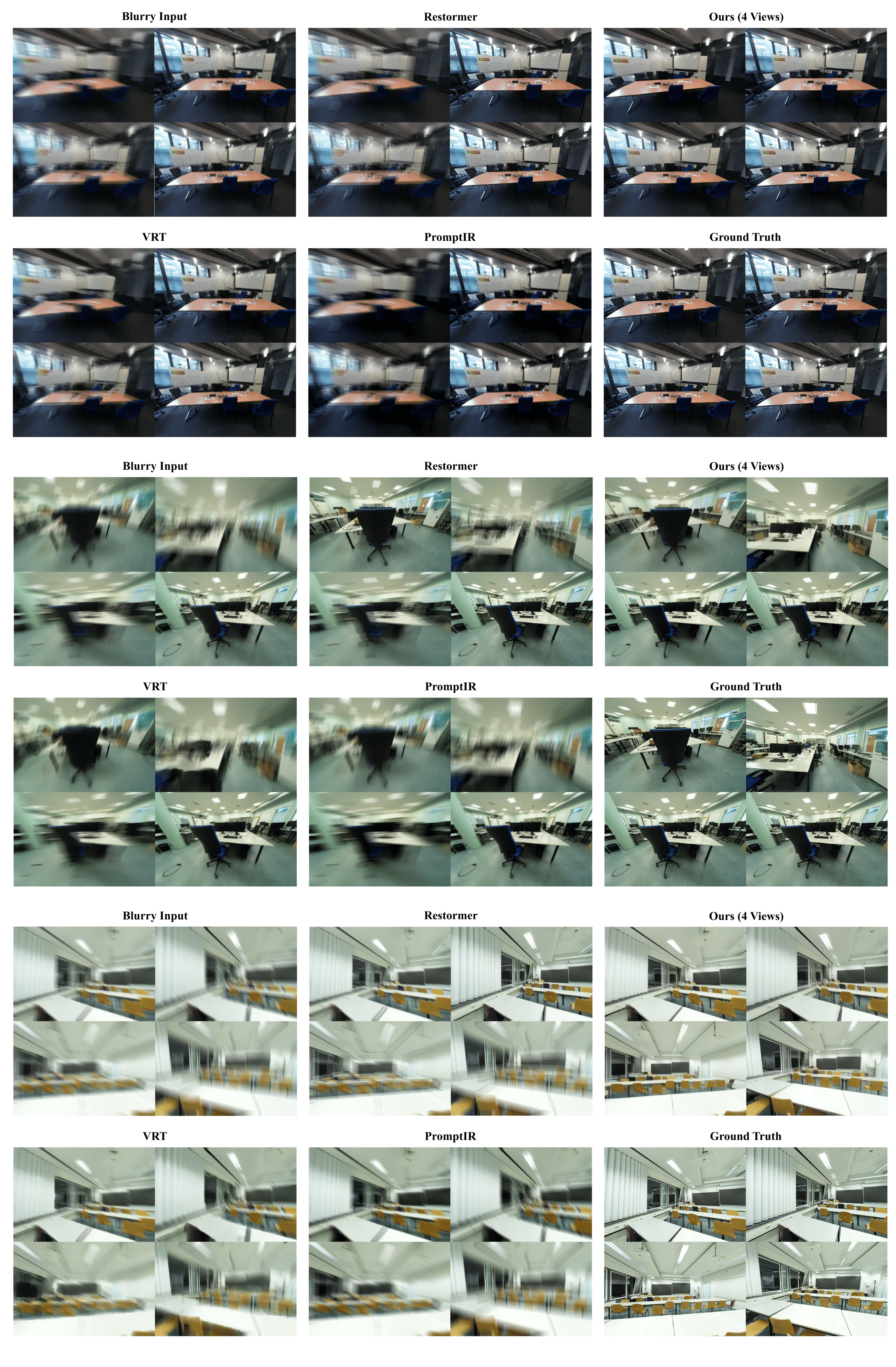}  
    \vspace{-0.2cm}
    \caption{
        \textbf{Additional Deblurring Results 1. } Our method uses all 4 input views to jointly denoise the images, performing significantly better than existing single-image-based methods \cite{zamir2022restormer,liang2024vrt,potlapalli2024promptir}.
    }
    \label{fig:supp_main_figure1}
\vspace{1.0cm}
\end{figure*}

\begin{figure*}[t]
    \centering
    \vspace{-0.2cm}
    \includegraphics[height=0.98\textheight]{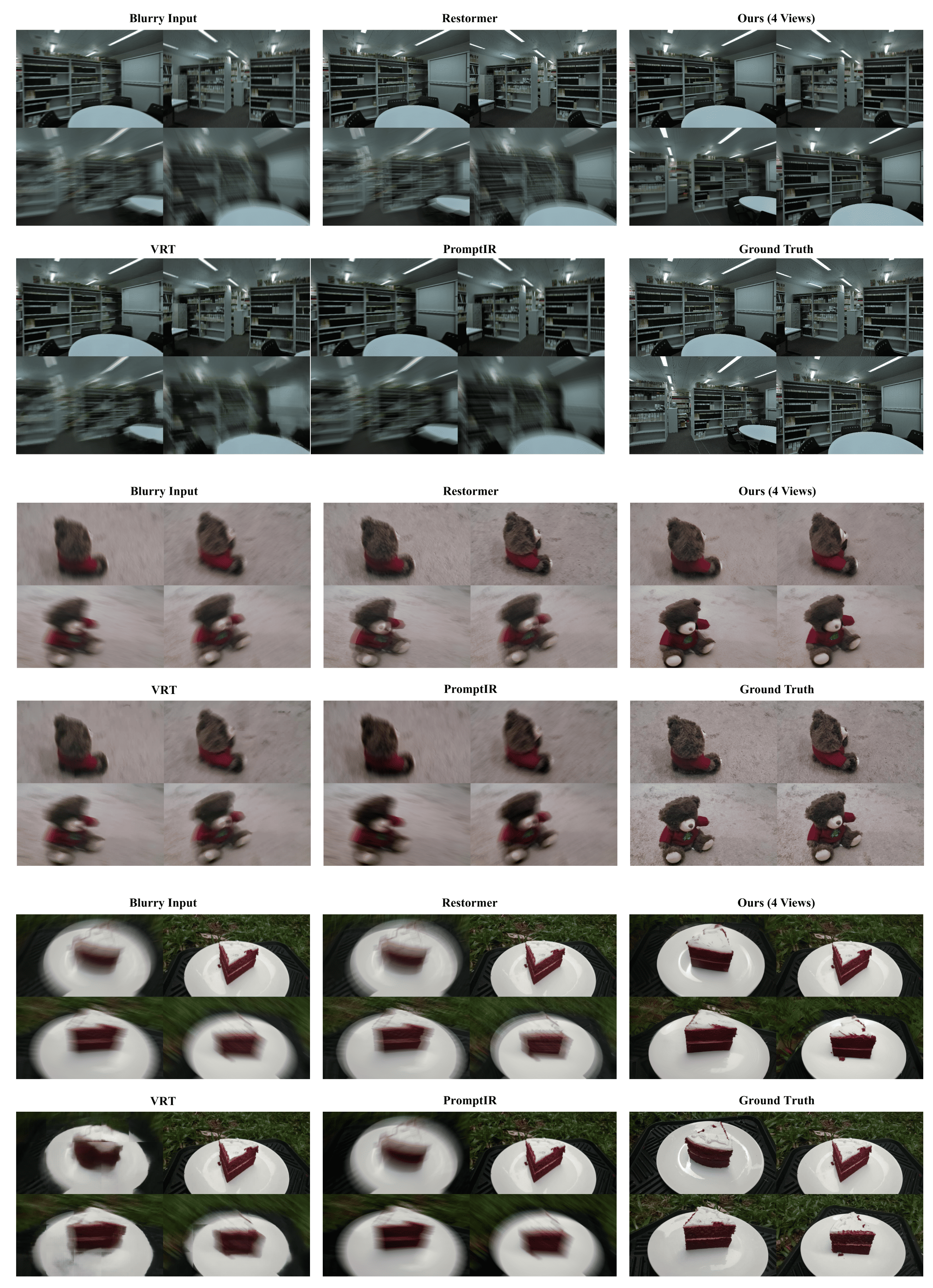}  
    \vspace{-0.2cm}
    \caption{
        \textbf{Additional Deblurring Results 2. } Our method uses all 4 input views to jointly denoise the images, performing significantly better than existing single-image-based methods \cite{zamir2022restormer,liang2024vrt,potlapalli2024promptir}.
    }
    \label{fig:supp_main_figure2}
\vspace{1.0cm}
\end{figure*}

\begin{figure*}[t]
    \centering
    \vspace{-0.5cm}
    \includegraphics[height=1.0\textheight]{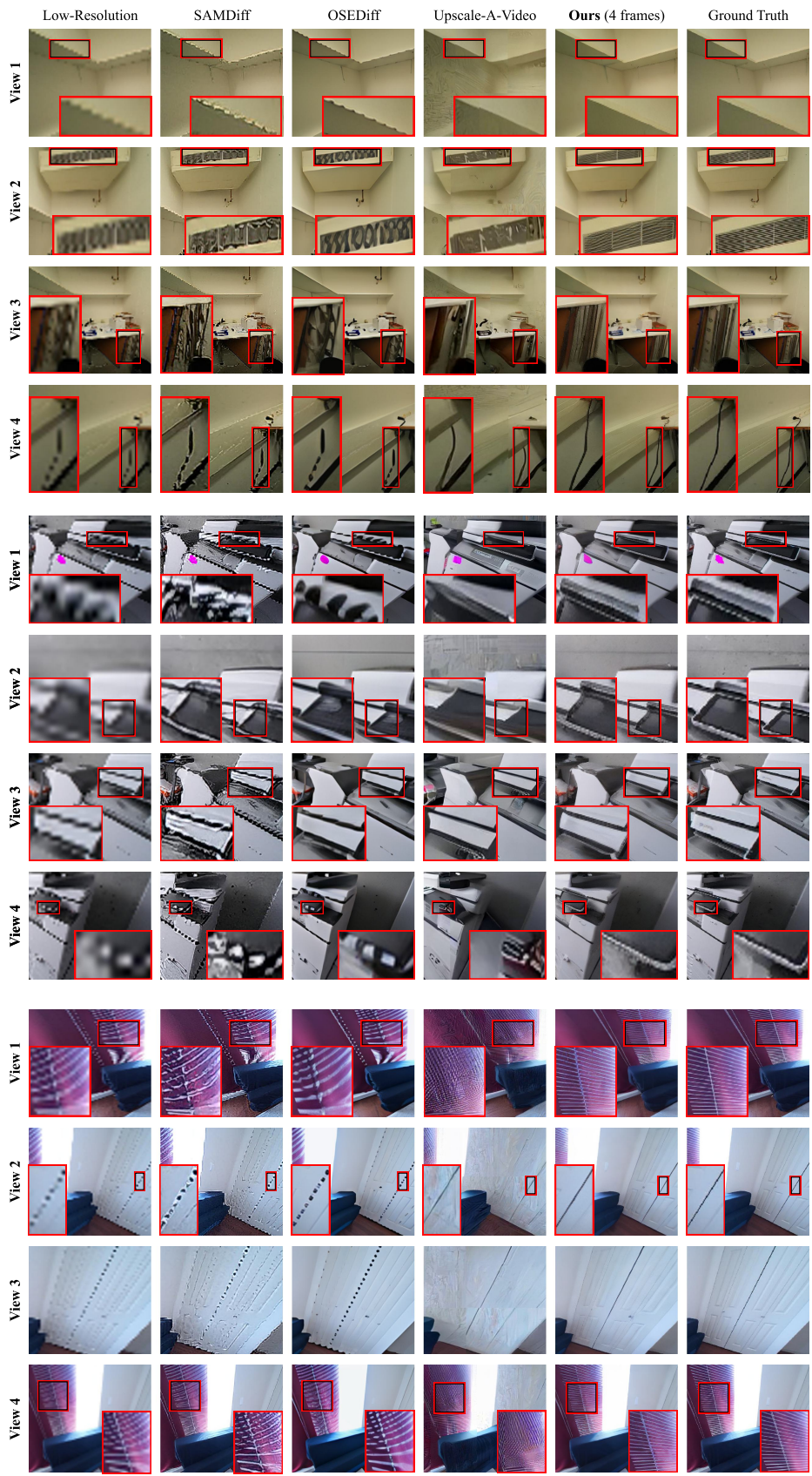}  
    \vspace{-0.3cm}
    \caption{
        \textbf{Additional Super-Resolution Comparison Results 1.} Our method uses all 4 input views to jointly denoise the images, performing significantly better than existing single-image-based methods~\cite{wang2024sam,wu2024one,zhou2024upscale}. Zoom in for the best view.
    }
    \label{fig:supp_SR_fig1}
\end{figure*}

\begin{figure*}[t]
    \centering

    \includegraphics[height=0.7\textheight]{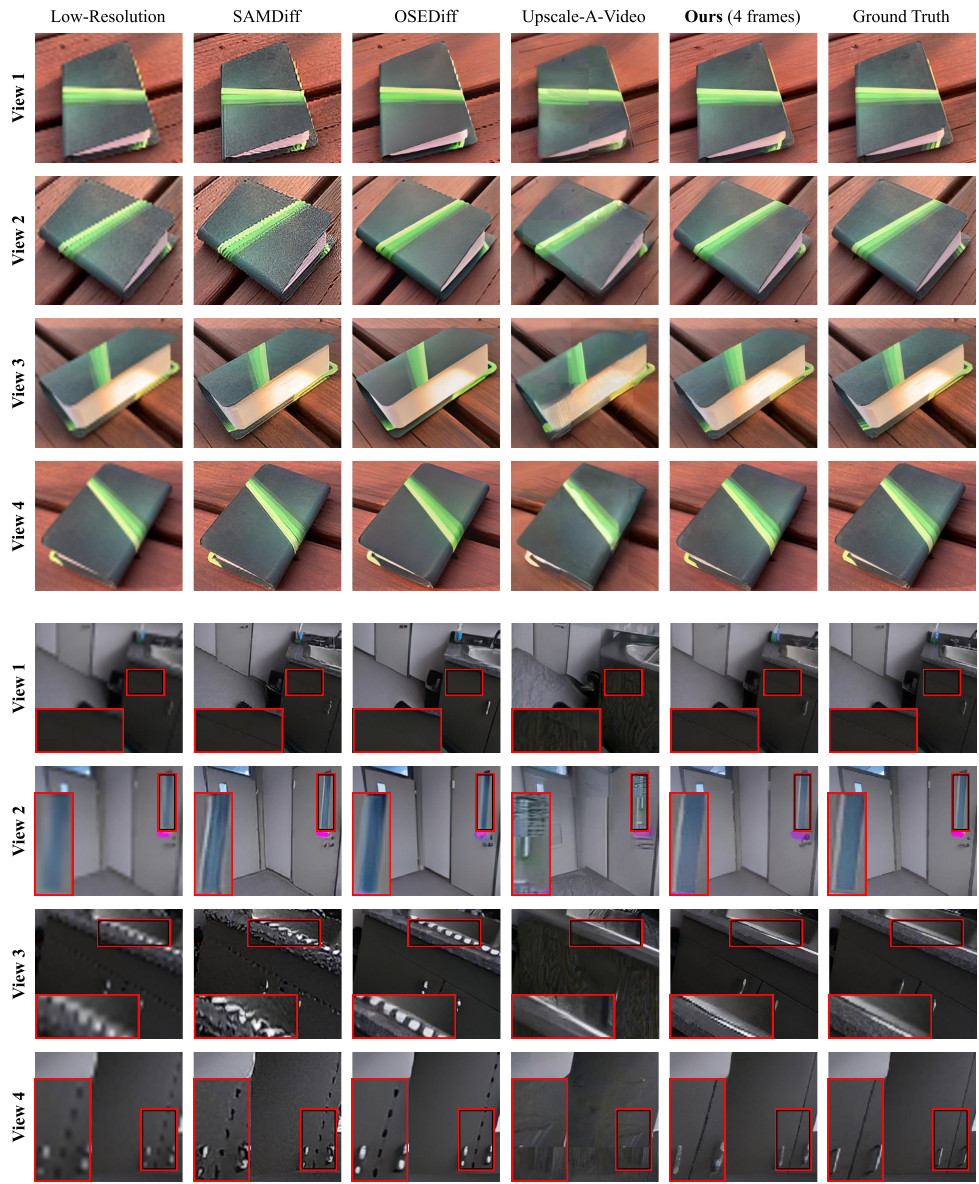}  
    \vspace{-0.3cm}
    \caption{
        \textbf{Additional Super Resolution Comparison Results 2.} Our method uses all 4 input views to jointly denoise the images, performing significantly better than existing single-image-based methods~\cite{wang2024sam,wu2024one,zhou2024upscale}. Zoom in for the best view.
    }
    \label{fig:supp_SR_fig2}
\vspace{1.0cm}
\end{figure*}

Fig.~\ref{fig:supp_main_figure1} and Fig.~\ref{fig:supp_main_figure2} show extra qualitative results of our \ours on motion deblurring. 
Fig.~\ref{fig:supp_SR_fig1} and Fig.~\ref{fig:supp_SR_fig2} show extra qualitative results of our \ours on super-resolution task.  
Our \ours show richer details in super-resolution and sharp deblurring restoration quality than other methods on both object level~\cite{reizenstein2021common} and scene level~\cite{yeshwanth2023scannet++} testing datasets.

\end{document}